\documentclass[10pt]{article}

\usepackage{graphicx}
\usepackage{color}

\usepackage[authoryear]{natbib}

\usepackage{bm}
\usepackage{amsmath}
\usepackage{amsfonts}
\usepackage{amsthm}
\usepackage{amssymb}
\usepackage{mathtools}
\mathtoolsset{showonlyrefs=true}
\newtheorem{theorem}{Theorem}

\DeclareMathOperator{\AIC}{AIC}
\DeclareMathOperator*{\argmin}{arg\,min}

\usepackage{booktabs}

\usepackage{url}
\usepackage{subcaption}
\usepackage{algorithm}
\usepackage{algorithmic}

\usepackage{multirow}

\usepackage[hidelinks]{hyperref}

\title{\vspace{-7mm}Statistical Test for Auto Feature Engineering\\ by Selective Inference}

\date{\today}

\makeatletter
\def\@fnsymbol#1{\ensuremath{\ifcase#1\or
{1}\or
{\ast}\or
{2}\or
{\dagger}\or
\else\@ctrerr\fi}}
\makeatother

\author{
Tatsuya Matsukawa\thanks{Nagoya University} \thanks{Equal contribution} ,
Tomohiro Shiraishi\footnotemark[1] \footnotemark[2] ,\\
Shuichi Nishino\footnotemark[1] ,
Teruyuki Katsuoka\footnotemark[1] ,
Ichiro Takeuchi\footnotemark[1] \thanks{RIKEN} \thanks{Corresponding author. e-mail: ichiro.takeuchi@mae.nagoya-u.ac.jp}
}
\begin{document}

\maketitle

\thispagestyle{empty}

\begin{abstract}
    \noindent
    Auto Feature Engineering (AFE) plays a crucial role in developing practical machine learning pipelines by automating the transformation of raw data into meaningful features that enhance model performance.
By generating features in a data-driven manner, AFE enables the discovery of important features that may not be apparent through human experience or intuition.
On the other hand, since AFE generates features based on data, there is a risk that these features may be overly adapted to the data, making it essential to assess their reliability appropriately.
Unfortunately, because most AFE problems are formulated as combinatorial search problems and solved by heuristic algorithms, it has been challenging to theoretically quantify the reliability of generated features.
To address this issue, we propose a new statistical test for generated features by AFE algorithms based on a framework called selective inference.
As a proof of concept, we consider a simple class of tree search-based heuristic AFE algorithms, and consider the problem of testing the generated features when they are used in a linear model.
The proposed test can quantify the statistical significance of the generated features in the form of $p$-values, enabling theoretically guaranteed control of the risk of false findings.

\end{abstract}

\newpage
\section{Introduction}
\label{sec:intro}
Feature engineering (FE), which is used to extract important features from raw data, plays a crucial role in many practical machine learning tasks.
While FE is often performed through trial and error based on experts' domain knowledge, data-driven approaches known as \emph{auto feature engineering (AFE)} are also promising and are being actively researched.
By generating features in a data-driven manner, AFE enables the discovery of important features that may not be apparent through human experience or intuition.
However, since AFE generates features based on data, there is a risk that these features may be overly adapted to the data, making it essential to appropriately assess their reliability.
In this study, we explore how to quantify the statistical reliability of features automatically generated by AFEs.

\begin{figure*}[t]
    \centering
    \includegraphics[width=0.75\linewidth]{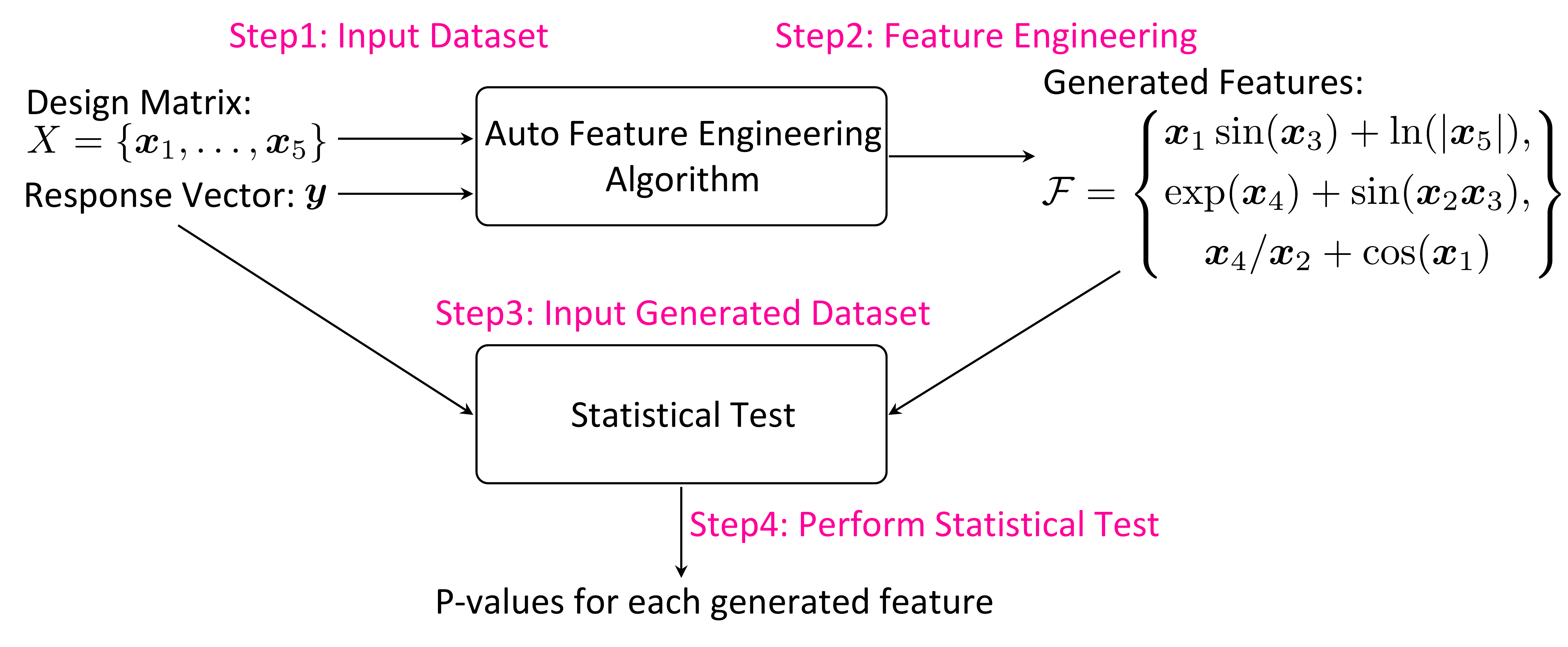}
    \caption{
        Overview of the proposed framework for evaluating the significance of features generated by the AFE algorithm.
        The input dataset $(X,\bm{y})$ is processed by an AFE algorithm to generate a set of features $\mathcal{F}$.
        The significance of each generated feature is then evaluated using a proposed selective inference method.
        Note that directly applying traditional statistical tests to generated features can lead to inflated type I error rates, as these features are data-dependent.
    }
    \label{fig:proposede_framework}
\end{figure*}

The AFE problem is essentially formulated as a combinatorial optimization problem because features are generated by recursively applying various operations to raw variables and previously generated features.
Figure~\ref{fig:proposede_framework} illustrates an example of feature engineering, where features such as $x_1 \sin(x_3) + \ln(|x_5|)$, $\exp(x_4) + \sin(x_2 x_3)$, and $x_4/x_2 + \cos(x_1)$ are generated by recursively applying various operations to the original raw variables $x_1, x_2, \ldots, x_5$, and previously generated features.
Therefore, most AFE algorithms are developed with the aim of finding approximate solutions to combinatorial search problems using heuristic search methods.
Since the properties of approximate solutions identified through such heuristic searches are difficult to characterize, quantifying the statistical uncertainties of the generated features is highly challenging.
As a result, to the best of our knowledge, no research has been conducted on statistical inference for the features generated by AFE algorithms.

To address this issue, we propose a new statistical test for automatically generated features by heuristic AFE algorithms, based on a framework called selective inference (SI).
SI is a statistical framework that allows for valid inference after data-dependent hypothesis selection procedures.
The basic idea of our proposed test is to conduct a statistical test conditional on the fact that the feature is generated by an AFE algorithm, which might otherwise lead to overfitting and inflated type I error rates.
By conditioning on the feature generation event, our proposed test adjusts for the bias introduced by the feature generation process, providing valid statistical inference for the generated features.
Our proposed test provides $p$-values for the automatically generated features, allowing users to assess their statistical significance and control the risk of falsely identifying irrelevant features.
This ensures that features passing the our proposed test, even those generated by heuristic AFE algorithms, can be confidently utilized in subsequent analyses.

In this paper, as a proof of concept, we consider a simple class of tree search-based heuristic AFE algorithms and address the problem of testing the generated features when they are used in a linear model.
We validate the effectiveness of our proposed method through experiments on both synthetic and real-world datasets.
The results demonstrate that our proposed test provides valid $p$-values for features generated by AFE algorithms.
Our implementation is available at \url{https://github.com/Takeuchi-Lab-SI-Group/statistical_test_for_auto_feature_engineering}.

\paragraph{Related Work.}
Various heuristic AFE methods have  been proposed in the literature.
For example, \emph{Data Science Machine} \citep{dsm_kanter2015deep} uses singular value decomposition to select generated features, while \emph{autofeat} \citep{autofeat_horn2020} performs feature generation based on $L_1$ regularization.
The limitation of all these heuristic strategies is that calculating all possible candidate features is computationally expensive.
To address this computational issue, many studies have adopted strategies that recursively generate more and more complex features \citep{autolearn_kaul2017autolearn, tfc_piramuthu2009iterative, safe_shi2020safe, openfe_zhang2022openfe, cognito_khurana2016cognito, ficus_markovitch2002feature}.
%
%
%
Additionally, methods using popular meta-heuristics such as genetic algorithms have also been explored \citep{ga_smith2003feature, automlga_shi2023automated, gaenergy_khan2020genetic}.
However, to the best of our knowledge, none of these approaches have focused on statistical reliability of generated features.
One common heuristic adopted by many AFE methods is known as the \emph{Expand-Reduce} strategy~\citep{dsm_kanter2015deep, autofeat_horn2020, obm_lam2021automated}.
This strategy consists of an \emph{expansion step}, where feature candidates are generated from the original raw variables or previously generated features, and a \emph{reduction step}, where features that contribute to the model's performance are selected.
In this study, as an example of commonly used heuristics in AFE, we address the problem of statistical testing for features generated by simple class of expand-reduce strategy.

SI has gained attention as a method of statistical inference for feature selection in linear models~\citep{taylor2015statistical, fithian2015selective}.
SI for various feature selection algorithms, such as marginal screening~\citep{lee2014exact}, stepwise FS~\citep{tibshirani2016exact}, and Lasso~\citep{lee2016exact}, has been explored and extended to more complex feature selection methods~\citep{yang2016selective, suzumura2017selective, hyun2018exact, rugamer2020inference, das2021fast}.
SI proves valuable not only for FS in linear models but also for statistical inference across various data-driven hypotheses, including unsupervised learning tasks such as OD~\citep{chen2020valid,tsukurimichi2021conditional}, segmentation~\citep{tanizaki2020computing,duy2022quantifying,le2024cad}, clustering~\citep{lee2015evaluating,gao2022selective}, change-point detection~\citep{duy2020computing,jewell2022testing}.
Moreover, SI is being applied not only to linear models but also to more complex models, such as kernel models~\citep{yamada2018post}, tree-structured models~\citep{neufeld2022tree}, Neural Networks~\citep{duy2022quantifying, miwa2023valid,shiraishi2024statistical}, and data analysis pipelines~\citep{shiraishi2024statistical2}.
However, existing SI studies have focused only on the inference of features selected from predefined candidates.
To our knowledge, there are no existing SI studies that perform inference on newly generated features as in AFE.
%
%
%
%

%
\newpage
\section{Preliminaries}
\label{sec:preliminaries}
Auto feature engineering (AFE) is the process of selecting, transforming, and creating relevant features from raw dataset to improve the performance of machine learning models.
In this study, we propose a framework for evaluating the significance of features generated by AFE algorithms using statistical hypothesis testing.
Figure~\ref{fig:proposede_framework} illustrates the overview of the proposed framework.
The AFE algorithm considered in this study will be discussed in detail in~\S\ref{sec:afe_algorithm}.
\paragraph{Problem Setup.}
Let us consider AFE for regression problem with $n$ instances and $m$ features.
We denote the dataset $(X, \bm{y})$, where $X=\left(\bm{x}_1,\bm{x}_2, \ldots, \bm{x}_m\right) \in \mathbb{R}^{n \times m}$ is the fixed design matrix and $\bm{y} \in \mathbb{R}^n$ is the response vector.
We assume that the observed response vector $\bm{y}$ is a random realization of the following random response vector
\begin{equation}
    \label{eq:statistical_model}
    \bm{Y} = \bm{\mu}(X) + \bm{\varepsilon},\ \bm{\varepsilon}\sim\mathcal{N}(\bm{0}, \Sigma)
\end{equation}
where $\bm{\mu}(X) \in \mathbb{R}^{n}$ is the unknown true value function, while $\bm{\varepsilon} \in \mathbb{R}^{n}$ is normally distributed noise with covariance matrix $\Sigma$ which is known or estimable from an independent dataset\footnote{
    We discuss the robustness of the proposed method when the covariance matrix is unknown and the noise deviates from the Gaussian distribution in Appendix~\ref{app:robustness}.
}.
Although we do not pose any functional form on the true value function $\bm{\mu}(X)$ for theoretical justification, we consider a scenario where the true values $\bm{\mu}(X)$ can be well approximated with a linear model by adding a feature set $\mathcal{F}\in \mathbb{R}^{n\times k}$, generated from the dataset $(X, \bm{y})$, to the design matrix $X$, where $k$ is the number of generated features.
In this case, the true value function can be approximated as $\bm{\mu}(X) \approx [X, \mathcal{F}]\bm{\beta}$ with $(m+k)$-dimensional coefficient vector $\bm{\beta}$.
This is a common setting in the field of SI, referred to as the \emph{saturated model} setting.
Using the above notations, a feature engineering algorithm is represented as a mapping:
\begin{equation}
    \label{eq:fe_algorithm_as_mapping}
    f_\mathrm{AFE}\colon \mathbb{R}^{n\times m}\times \mathbb{R}^{n} \ni (X, \bm{y})
    \mapsto \mathcal{F} \in \mathbb{R}^{n\times \mathbb{N}},
\end{equation}
whose specific procedure are detailed in~\S\ref{sec:afe_algorithm}.
\paragraph{Statistical Test for Generated Features.}
Given the generated features by the AFE algorithm in~\eqref{eq:fe_algorithm_as_mapping}, the statistical significance of the generated features can be quantified based on the coefficients of the linear model fitted with the concatenated design matrix $[X, \mathcal{F}]$.
To formalize this, we denote the design matrix after adding the generated features as $X_{\mathcal{F}} = [X, \mathcal{F}] \in \mathbb{R}^{n \times (m+k)}$ for simplicity.
Using these notations, the least-squares solution of the linear model after adding the generated features is expressed as
\begin{equation}
    \label{eq:betahat}
    \hat{\bm{\beta}}
    =
    \left(
    X_{\mathcal{F}}^\top X_{\mathcal{F}}
    \right)^{-1}
    X_{\mathcal{F}}^\top
    \bm{y}.
\end{equation}
Similarly, we consider the population least-square solution for the unobservable true value vector $\bm{\mu}(X)$ in \eqref{eq:statistical_model}, which is defined as
\begin{equation}
    \label{eq:beta_star}
    \bm{\beta}^\ast
    =
    \left(
    X_{\mathcal{F}}^\top X_{\mathcal{F}}
    \right)^{-1}
    X_{\mathcal{F}}^\top
    \bm{\mu}(X).
\end{equation}
To quantify the statistical significance of the generated features, we consider the following null hypothesis and the alternative hypothesis:
\begin{equation}
    \label{eq:test}
    \mathrm{H}_{0,j}\colon \beta^\ast_j = 0\
    \text{v.s.}\
    \mathrm{H}_{1,j}: \beta^\ast_j \neq 0,\
    j \in \{m+1,\ldots,m+k\}.
\end{equation}
In this testing problem, to test the coefficient of the generated feature itself, we take $j$ from $\{m+1,\ldots,m+k\}$.
Naturally, if we let $j \in [m]$, it becomes a test of the coefficients of the original features in the linear model with the generated features added, and all of the subsequent discussion can be carried out in exactly the same way.
Then, we will proceed with the discussion assuming $j \in [m+k]$.
\paragraph{Selective Inference (SI).}
For the statistical test in~\eqref{eq:test}, it is reasonable to use $\hat{\beta}_j,\ j\in[m+k]$ as the test statistic.
An important point when addressing this statistical test within the SI approach is that the test statistic is represented as a linear function of the observed response vector as
$\hat{\beta}_j = \bm{\eta}_j^\top \bm{y}, j \in [m+k]$,
where $\bm{\eta}_j \in \mathbb{R}^{n},\ j\in[m+k]$ is a vector that depends on $\bm{y}$ only through the generated features $\mathcal{F}$.
In SI, this property is utilized to perform statistical inference based on the sampling distribution of the test statistic conditional on $\mathcal{F}$.
More specifically, since $\bm{y}$ follows a normal distribution, it can be derived that the sampling distribution of the test statistic $\hat{\beta}_j = \bm{\eta}_j^\top \bm{y},\ j\in[m+k]$ conditional on $\mathcal{F}$ and the sufficient statistic of the nuisance parameters follows a truncated normal distribution.
By computing $p$-values based on this conditional sampling distribution represented as a truncated normal distribution, it is ensured that the type I error can be controlled even in finite samples.
For more details on SI, please refer to the explanations in~\S\ref{sec:selective_inference} or literatures such as \citet{taylor2015statistical,fithian2015selective,lee2014exact}.
\newpage
\section{Auto Feature Engineering}
\label{sec:afe_algorithm}
In this section, we describe the AFE algorithm $f_\mathrm{AFE}$ in~\eqref{eq:fe_algorithm_as_mapping} used in this study.
Note that the novelty of this study is to propose a method for quantify the statistical significance of the generated features by the AFE algorithm, not the AFE algorithm itself.
\subsection{Directed Tree Search Algorithm}
As a simple AFE algorithm, we consider a procedure that repeatedly applies a given set of transformations $\mathcal{T}=\{t_1,\ldots,t_{|\mathcal{T}|}\}$ to the original features $\{\bm{x}_1, \ldots, \bm{x}_m\}$ and selects the resulting features based on their fitness to the response vector $\bm{y}$.
Examples of such transformations include elementary functions such as sine, cosine, and exponential, arithmetic operations such as multiplication and division.

\begin{figure*}[h]
    \centering
    \includegraphics[width=0.77\linewidth]{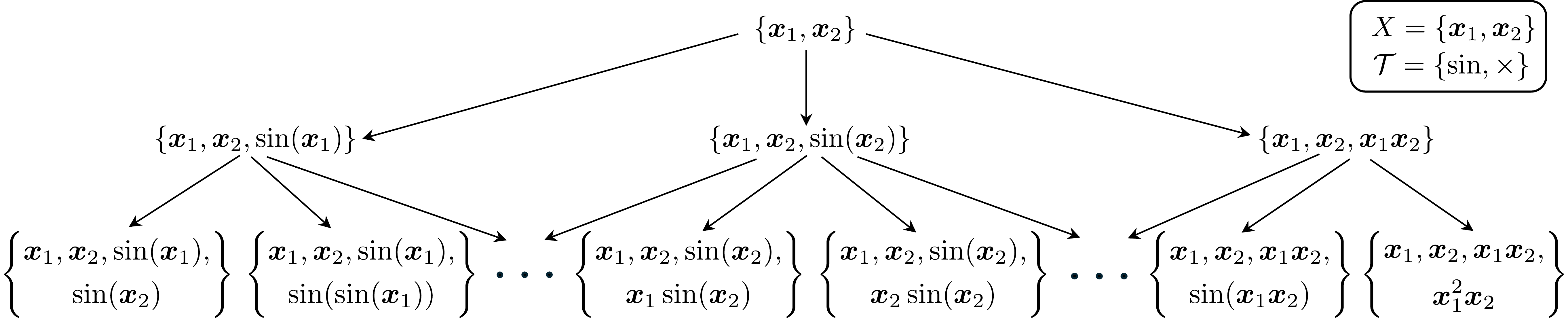}
    \caption{Schematic illustration of the feature generation process as a directed tree search.}
    \label{fig:feature_generation_tree}
\end{figure*}

To formalize this AFE algorithm, we employ a directed tree structure to represent the process.
In this structure, each node represents a set of features (with the root node corresponding to the original set of features $\{\bm{x}_1,\ldots,\bm{x}_m\}$), and a directed edge signifies the addition of a new feature generated by applying a transformation from $\mathcal{T}$ to one feature in the parent node.
Thus, the AFE algorithm can be regarded as a search within this directed tree for a node containing an appropriate set of features, as evaluated by their fitness to the response vector $\bm{y}$.
A schematic illustration of the feature generation process is shown in Figure~\ref{fig:feature_generation_tree}.
\subsection{Details of AFE Algorithm}
\begin{figure}[htbp]
    \centering
    \includegraphics[width=1.0\linewidth]{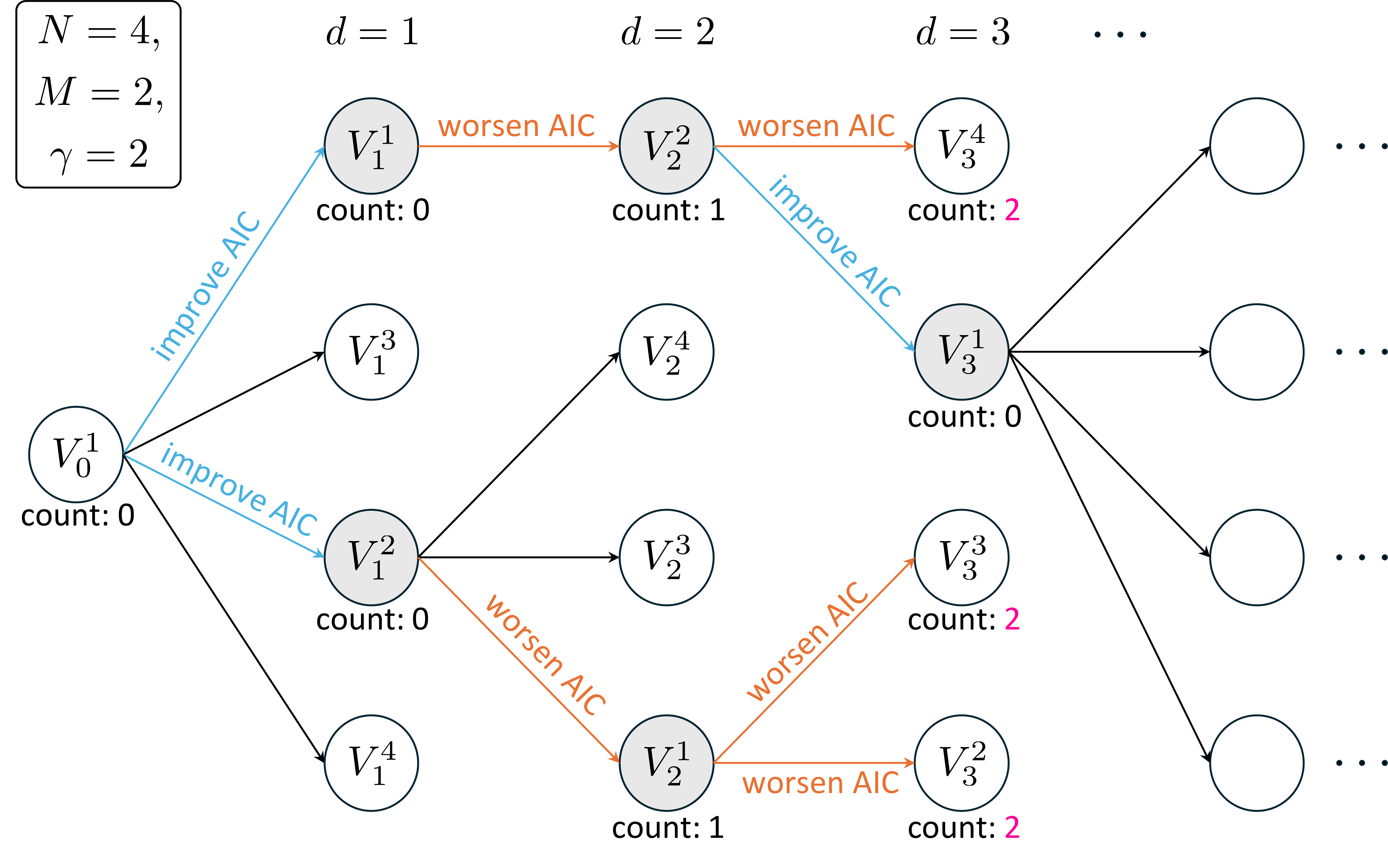}
    \caption{
        Schematic illustration of the AFE algorithm.
        The search proceeds roughly as follows: for each increase in depth, $N(=4)$ nodes are randomly generated (each node has a feature set that adds one new feature to the parent node).
        Here, $V_i^j$ denotes a node, with the subscript $i$ denoting the depth of the node and the superscript $j$ denoting the rank of the node within the same depth, ordered by AIC.
        Then, $M(=2)$ nodes with the best AIC among the $N$ generated nodes are selected as parents of the next depth nodes (nodes filled in gray).
        This operation is repeated until the maximum depth $D$ is reached.
        %
        Furthermore, to handle exceptions, we check whether the AIC has improved compared to the best AIC at the previous depth and record the number of consecutive times the AIC has not improved for each node (note that it is reset to $0$ at $V_3^1$).
        The AFE algorithm prevents the node with this counter value greater than or equal to $\gamma(=2)$ from becoming parent node.
        In this figure, at depth $3$, all nodes except $V_3^1$ cannot become parents, then the next feature generation is performed only from $V_3^1$.
    }
    \label{fig:feature_generation_algorithm}
\end{figure}

The search space of the directed tree, as described above, is exponentially large, making it computationally infeasible to explore all possible nodes exhaustively.
%
To address this issue, we propose a simple directed tree search algorithm governed by four hyperparameters: the maximum depth $D$, the maximum number of nodes to be generated $N$ at each depth, the maximum number of nodes used for generation $M$ at each depth, and the tolerance parameter $\gamma$ for cases where increasing the depth does not lead to an improvement in fitness.
In this study, we simply employ AIC as a measure of fitness to the response vector $\bm{y}$.
A schematic illustration of the AFE algorithm as a directed tree search is shown in Figure~\ref{fig:feature_generation_algorithm}, and the overall algorithm is summarized in Algorithm~\ref{alg:overall_tree_search}.

The AFE algorithm, evident from its procedure, is significantly influenced by its hyperparameters.
%
For instance,
%
1) setting large $D$ and $\gamma$, a small $N$, and $M=1$ results in the generation of a few parent node candidates and exploration deep into the graph without fear of AIC deterioration, which is similar to the behavior of depth-first search.
2) setting small $D$ and $\gamma$, large $N$ and $M$ leads to exploring a greater number of potential nodes at each depth without going too deep, which is similar to the behavior of breadth-first search.
%
%
%
\begin{algorithm}[htbp]
    \caption{Overall of the AFE Algorithm}
    \label{alg:overall_tree_search}
    \begin{algorithmic}[1]
        \REQUIRE Set of original features $X=\{\bm{x}_1, \ldots, \bm{x}_m\}$, response vector $\bm{y}$.
        \STATE Initialize the set of nodes at depth $0$ as $\mathcal{V}_0 = \{X\}$
        \STATE Cache the best AIC at depth $0$ as $\mathtt{best\_aic[0]}=\AIC(X, \bm{y})$
        \STATE Cache the consecutive non-improvement count for node $X$ as $\mathtt{no\_improve[X]}=0$
        \FOR{each depth $d \in \{0, \ldots, D-1\}$}
        \STATE Initialize the set of nodes at depth $d+1$ as $\mathcal{V}_{d+1} = \emptyset$
        \WHILE{$|\mathcal{V}_{d+1}| < N$ and new feature $\bm{f}$ can be generated from $\mathcal{V}_d$ and $\mathcal{T}$}
        \STATE Randomly select a node $V \in \mathcal{V}_d$ to expand
        \STATE Generate new feature $\bm{f}=t(\bm{x})$, where $(t, \bm{x})$ are randomly selected from $\mathcal{T}\times V$
        \IF{feature $\bm{f}$ do not exhibit multicollinearity with $V$}
        \STATE Add the node representing feature set $V \cup \{\bm{f}\}$ to $\mathcal{V}_{d+1}$
        \ENDIF
        \ENDWHILE
        \STATE Reduce nodes as $\mathcal{V}_{d+1}=\texttt{ReducingNode}(\mathcal{V}_{d+1})$ based on the AIC (Algorithm~\ref{alg:reducing_node})
        \IF{$\mathcal{V}_{d+1}$ is empty}
        \STATE Decrement the depth $d$ and break the for loop
        \ENDIF
        \ENDFOR
        \ENSURE Generated features in node with the best AIC in $\mathcal{V}_{d+1}$, i.e., $\argmin_{V \in \mathcal{V}_{d+1}} \AIC(V, \bm{y})\setminus X$
    \end{algorithmic}
\end{algorithm}
\begin{algorithm}[htbp]
    \caption{Reducing Node}
    \label{alg:reducing_node}
    \begin{algorithmic}[1]
        \REQUIRE Set of nodes $\mathcal{V}_{d+1}$ at depth $d+1$
        \STATE Number each node of $\mathcal{V}_{d+1}$ in ascending order of AIC as $\mathcal{V}_{d+1}=\{V_{d+1}^{1}, \ldots, V_{d+1}^{|\mathcal{V}_{d+1}|}\}$ 
        \STATE Cache the best AIC at depth $d+1$ as $\mathtt{best\_aic[d+1]}=\AIC(V_{d+1}^1, \bm{y})$
        \STATE Initialize set of nodes to be returned as $\mathcal{V}_{d+1}^\prime = \emptyset$
        \FOR{each node $V \in \mathcal{V}_{d+1}$}
        \STATE Cache count $\mathtt{no\_improve[V]}$ as $0$ if $\AIC(V, \bm{y}) < \mathtt{best\_aic[d]}$, else $\mathtt{no\_improve[pa(V)]} + 1$
        \IF{$\mathtt{no\_improve[V]} < \gamma$}
        \STATE Add node $V$ to $\mathcal{V}_{d+1}^\prime$
        \ENDIF
        \IF{$|\mathcal{V}_{d+1}^\prime| = M$}
        \STATE Break the for loop
        \ENDIF
        \ENDFOR
        \ENSURE $\mathcal{V}_{d+1}^\prime$
    \end{algorithmic}
\end{algorithm}

\newpage
\section{Selective Inference for Generated Features}
\label{sec:selective_inference}
To perform statistical hypothesis testing for generated features, it is necessary to consider how the data influenced the finally generated features through the whole process of the AFE algorithm.
We address this challenge by utilizing the SI framework.
In the SI framework, statistical test is performed based on the sampling distribution conditional on the process by which the features are generated from the data\footnote{
    This can also be interpreted as the process by which the data selects the generated features from the set of all potentially generated features.
    In the context of SI, this interpretation as a selection from the range of the algorithm based on the data is more natural.
}, thereby incorporating the influence of how data is used to generate the features.
%
\subsection{Concept of Selective Inference}
In SI, $p$-values are computed based on the null distribution conditional on an event that a certain hypothesis is generated.
The goal of SI is to compute a $p$-value such that
\begin{equation}
    \label{eq:conditional_type_i_error_rate}
    \mathbb{P}_{\mathrm{H}_0}
    \left(
    p\leq \alpha \mid
    \mathcal{F}_{\bm{Y}} = \mathcal{F}
    \right)
    = \alpha,\ \forall\alpha\in (0,1),
\end{equation}
where $\mathcal{F}_{\bm{Y}}$ indicates the random set of generated features given the random response vector $\bm{Y}$ and the fixed design matrix $X$, thereby making the $p$-value is a random variable.
Here, the condition part $\mathcal{F}_{\bm{Y}} = \mathcal{F}$ in~\eqref{eq:conditional_type_i_error_rate} indicates that we only consider response vectors $\bm{Y}$ from which a certain set of features are generated.
If the conditional type I error rate can be controlled as in~\eqref{eq:conditional_type_i_error_rate} for all possible hypotheses $\mathcal{F}\in \mathcal{F}(X)$ where $\mathcal{F}(X)$ represents the set of all possible generated features from the fixed design matrix $X$, then, by the law of total probability, the marginal type I error rate can also be controlled for all $\alpha\in (0,1)$ because
\begin{equation}
    \label{eq:marginal_type_i_error_rate}
    \mathbb{P}_{\mathrm{H}_0}(p\leq \alpha) =
    \sum_{\mathcal{F}\in \mathcal{F}(X)}
    \mathbb{P}_{\mathrm{H}_0}(\mathcal{F})
    \mathbb{P}_{\mathrm{H}_0}
    \left(
    p\leq \alpha \mid \mathcal{F}_{\bm{Y}} = \mathcal{F}
    \right)
    = \alpha.
\end{equation}
Therefore, in order to perform valid statistical test, we can employ $p$-values conditional on the hypothesis selection event.
To compute a $p$-value that satisfies~\eqref{eq:conditional_type_i_error_rate}, we need
to derive the sampling distribution of the test statistic
\begin{equation}
    \label{eq:conditional_test_statistic}
    \bm{\eta}_j^\top\bm{Y}
    \mid
    \{
    \mathcal{F}_{\bm{Y}} = \mathcal{F}_{\bm{y}}
    \}.
\end{equation}
%
\subsection{Selective $p$-value}
To perform statistical test based on the conditional sampling distribution in~\eqref{eq:conditional_test_statistic}, we introduce an additional condition on the sufficient statistic of the nuisance parameter $\mathcal{Q}_{\bm{Y}}$, defined as
\begin{equation}
    \label{eq:nuisance_parameter}
    \mathcal{Q}_{\bm{Y}} =
    \left(
    I_{n} - \frac{\Sigma\bm{\eta}\bm{\eta}^\top}{\bm{\eta}^\top\Sigma\bm{\eta}}
    \right)
    \bm{Y}.
\end{equation}
This additional conditioning on $\mathcal{Q}_{\bm{Y}}$ is a standard practice in the SI literature required for computational tractability\footnote{
The nuisance component $\mathcal{Q}_{\bm{Y}}$ corresponds to the component $\bm{z}$ in the seminal paper~\citet{lee2016exact} (see Sec. 5, Eq. (5.2), and Theorem 5.2) and is used in almost all the SI-related works that we cited.
}.
Based on the additional conditioning on $\mathcal{Q}_{\bm{Y}}$, the following theorem tells that the conditional $p$-value that satisfies~\eqref{eq:conditional_type_i_error_rate} can be derived by using a truncated normal distribution.
\begin{theorem}
    \label{thm:conditional_sampling_distribution}
    Consider a fixed design matrix $X$, a random response vector $\bm{Y}\sim\mathcal{N}(\bm{\mu}, \Sigma)$ and an observed response vector $\bm{y}$.
    Let $\mathcal{F}_{\bm{Y}}$ and $\mathcal{F}_{\bm{y}}$ be the obtained set of generated features, by applying an auto feature engineering algorithm in the form of~\eqref{eq:fe_algorithm_as_mapping} to $(X, \bm{Y})$ and $(X, \bm{y})$, respectively.
    Let $\bm{\eta}\in \mathbb{R}^{n}$ be a vector depending on $\mathcal{F}_{\bm{y}}$, and consider a test statistic in the form of $T(\bm{Y})=\bm{\eta}^\top \bm{Y}$.
    Furthermore, define the nuisance parameter $\mathcal{Q}_{\bm{Y}}$ as in~\eqref{eq:nuisance_parameter}.

    Then, the conditional distribution
    \begin{equation}
        T(\bm{Y}) \mid
        \{
        \mathcal{F}_{\bm{Y}} = \mathcal{F}_{\bm{y}},
        \mathcal{Q}_{\bm{Y}} = \mathcal{Q}_{\bm{y}}
        \}
    \end{equation}
    is a truncated normal distribution $\mathrm{TN}(\bm{\eta}^\top\bm{\mu}, \bm{\eta}^\top\Sigma\bm{\eta}, \mathcal{Z})$ with the mean $\bm{\eta}^\top\bm{\mu}$, the variance $\bm{\eta}^\top\Sigma\bm{\eta}$, and the truncation intervals $\mathcal{Z}$.
    The truncation intervals $\mathcal{Z}$ is defined as
    \begin{equation}
        \label{eq:trancatio_intervals}
        \mathcal{Z} = \left\{
        z\in \mathbb{R}\mid
        \mathcal{F}_{\bm{a}+\bm{b}z} = \mathcal{F}_{\bm{y}},
        \right\},\
        \bm{a} = \mathcal{Q}_{\bm{y}},\
        \bm{b} = \bm{\Sigma\eta}/\bm{\eta}^\top\Sigma\bm{\eta}.
    \end{equation}
\end{theorem}
The proof of Theorem~\ref{thm:conditional_sampling_distribution} is deferred to Appendix~\ref{app:proof_truncated}.
By using the sampling distribution of the test statistic $T(\bm{Y})$ conditional on $\mathcal{F}_{\bm{Y}} = \mathcal{F}_{\bm{y}}$ and $\mathcal{Q}_{\bm{Y}} = \mathcal{Q}_{\bm{y}}$ in Theorem~\ref{thm:conditional_sampling_distribution}, we can define the selective $p$-value as
\begin{equation}
    \label{eq:selective_p_value}
    p_\mathrm{selective} =
    \mathbb{P}_{\mathrm{H}_0}
    \left(
    |T(\bm{Y})| \geq |T(\bm{y})| \mid
    \mathcal{F}_{\bm{Y}} = \mathcal{F}_{\bm{y}},
    \mathcal{Q}_{\bm{Y}} = \mathcal{Q}_{\bm{y}}
    \right).
\end{equation}
\begin{theorem}
    \label{thm:property_of_selective_p_value}
    The selective $p$-value defined in~\eqref{eq:selective_p_value} satisfies the property in~\eqref{eq:conditional_type_i_error_rate}, i.e.,
    \begin{equation}
        \label{eq:thm_first_term}
        \mathbb{P}_{\mathrm{H}_0}
        \left(
        p_\mathrm{selective} \leq \alpha \mid
        \mathcal{F}_{\bm{Y}} = \mathcal{F}_{\bm{y}}
        \right)
        = \alpha,\ \forall\alpha\in (0,1).
    \end{equation}
    Then, the selective $p$-value also satisfies the following property of a valid $p$-value:
    \begin{equation}
        \label{eq:thm_second_term}
        \mathbb{P}_{\mathrm{H}_0}(p_\mathrm{selective} \leq \alpha) = \alpha,\ \forall\alpha\in (0,1).
    \end{equation}
\end{theorem}
%
%
%
The proof of Theorem~\ref{thm:property_of_selective_p_value} is deferred to Appendix~\ref{app:proof_property_of_selective_p_value}.
This theorem guarantees that the selective $p$-value is uniformly distributed under the null hypothesis $\mathrm{H}_0$, and thus can be used to conduct the valid statistical inference in~\eqref{eq:test}. 
Once the truncation intervals $\mathcal{Z}$ is identified, the selective $p$-value in~\eqref{eq:selective_p_value} can be easily computed by Theorem~\ref{thm:conditional_sampling_distribution}.
Thus, the remaining task is reduced to identifying the truncation intervals $\mathcal{Z}$.

\newpage
\section{Computations of Truncation Intervals}
\label{sec:computation_truncation_intervals}
From the discussion in~\S\ref{sec:selective_inference}, it is suffice to identify the one-dimensional subset $\mathcal{Z}$ in~\eqref{eq:trancatio_intervals} to conduct the statistical test.
In this section, we propose a novel line search method to efficiently identify the $\mathcal{Z}$.
\subsection{Overview of the Line Search}
The difficulty in identifying the $\mathcal{Z}$ arises from the exponential number of possible paths that the AFE algorithm can traverse in the tree structure during feature generation.
To address this challenge, we propose an efficient search method based on parametric programming.
The core idea is to compute the interval within which the nodes traversed by the AFE algorithm remain unchanged in the tree structure, ensuring that the same set of features is generated.

\begin{figure*}[t]
    \centering
    \includegraphics[width=0.9\linewidth]{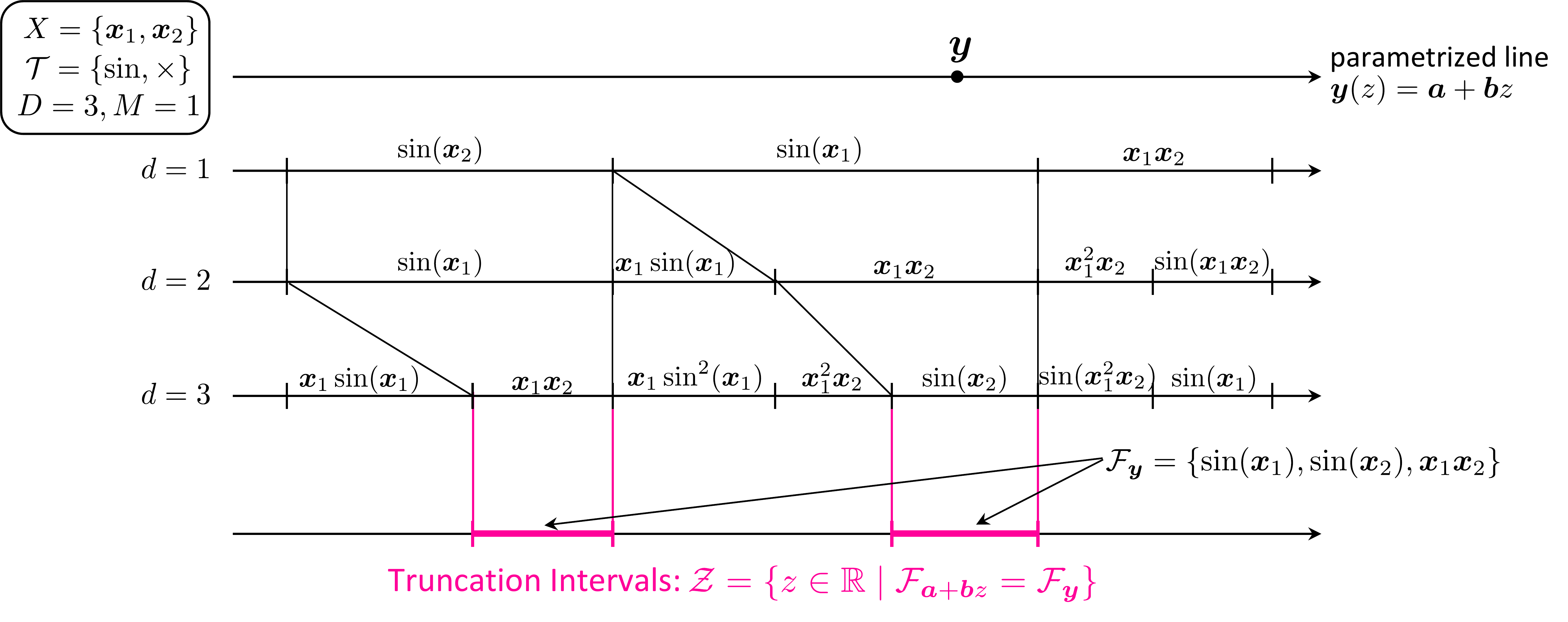}
    \caption{
        Schematic illustration of the proposed line search method to identify the truncation intervals $\mathcal{Z}$.
        We first compute the interval within which the generated features by the AFE algorithm remain unchanged.
        Then, we identify the truncation intervals $\mathcal{Z}$ by taking the union of some intervals based on parametric-programming.
    }
    \label{fig:demo_pp}
\end{figure*}

In the following, we present an overview of the proposed line search method to identify the truncation intervals $\mathcal{Z}$ using parametric programming, leveraging this interval computation.
Subsequently, we will elaborate on the details of this interval computation procedure.
An overview of the proposed line search method is illustrated in Figure~\ref{fig:demo_pp}.
\subsection{Parametric-Programming}
\label{subsec:parametric_programming}
To identify the truncation intervals $\mathcal{Z}$, we assume that we have a procedure to compute the interval $[L_z, U_z]$ for any $z\in\mathbb{R}$ which satisfies: for any $r\in [L_z, U_z]$, the outputs of the AFE algorithm from $(X, \bm{a}+\bm{b}r)$ and $(X, \bm{a}+\bm{b}z)$ are the same, i.e.,
\begin{equation}
    \forall r\in [L_z, U_z],\
    \mathcal{F}_{\bm{a}+\bm{b}r}=\mathcal{F}_{\bm{a}+\bm{b}z}.
\end{equation}
Then, the truncation intervals $\mathcal{Z}$ can be obtained by the union of the intervals $[L_z, U_z]$ as
\begin{equation}
    \label{eq:pp}
    \mathcal{Z} =
    \bigcup_{
    z\in\mathbb{R}\mid
    \mathcal{F}_{\bm{a}+\bm{b}z}=\mathcal{F}_{\bm{y}}
    }
    [L_z, U_z].
\end{equation}
The procedure in~\eqref{eq:pp} is commonly referred to as parametric-programming (e.g., lasso regularization path).
We discuss the details of the procedure to compute the interval $[L_z, U_z]$ by considering the nodes traversed by the AFE algorithm.
\subsection{Interval Computation}
\label{subsec:computation_interval}
We subsequently elaborate on the computational procedure for obtaining the interval $[L_z, U_z]$ for any $z\in\mathbb{R}$, as referenced in \S\ref{subsec:parametric_programming}.
To compute $[L_z, U_z]$, it is sufficient to consider the conditions under which applying the AFE algorithm to $(X, \bm{a}+\bm{b}r)$ leads to traversing the exact same nodes as when applied to $(X, \bm{a}+\bm{b}z)$, irrespective of the response vector $\bm{a}+\bm{b}r$.
Note that, to ensure determinism in the AFE algorithm, we adopt the convention of fixing the random seed to a constant value at the algorithm's initiation.

Upon reviewing Algorithms~\ref{alg:overall_tree_search} and~\ref{alg:reducing_node}, it becomes apparent that their procedures depend on the response vector solely at two specific points within Algorithm~\ref{alg:reducing_node}: the AIC-based sorting in the first line and the AIC improvement assessment in the fifth line.
Thus, we can deduce that the nodes traversed by the AFE algorithm remain unchanged if all AIC comparisons maintain the same order relations as when the algorithm is applied to $(X, \bm{a}+\bm{b}z)$.
This condition can be readily translated into multiple inequalities, each taking the form $\{r\in\mathbb{R}\mid \AIC(V_i, \bm{a}+\bm{b}r)\leq \AIC(V_j, \bm{a}+\bm{b}r)\}$, where $V_i$ and $V_j$ represent the sets of features.

Finally, our objective is reduced to solving the inequality $\AIC(V_i, \bm{a}+\bm{b}r)\leq \AIC(V_j, \bm{a}+\bm{b}r)$ for $r$.
Here, the AIC value of $(V, \bm{a}+\bm{b}r)$ can be expressed as:
\begin{equation}
    \AIC(V, \bm{a}+\bm{b}r) =
    (\bm{a}+\bm{b}r)^\top
    \Lambda
    (\bm{a}+\bm{b}r) + 2|V|,
\end{equation}
yielding a quadratic function of $r$, where $\Lambda=\Sigma^{-1} - \Sigma^{-1}V(V^\top\Sigma^{-1}V)^{-1}V^\top\Sigma^{-1}$.
Therefore, the condition is represented by the intersection of multiple quadratic inequalities, which can be analytically solved to obtain a union of multiple intervals.
For notational simplicity, we denote only the interval containing $z$ as $[L_z, U_z]$.

\newpage
\section{Numerical Experiments}
\label{sec:experiments}
\paragraph{Methods for Comparison.}
%
In our experiments, we compare the proposed method (\texttt{proposed}) with over-conditioning (\texttt{oc}) method (simple extension of SI literature to our setting), naive test (\texttt{naive}), Bonferroni correction (\texttt{bonferroni}), and data splitting (\texttt{ds}), in terms of type I error rate and power.
See Appendix~\ref{app:methods_for_comparison} for more details on the methods for comparison.
\paragraph{Experimental Setup.}
In all experiments, we set the significance level $\alpha=0.05$.
For the configuration of the AFE algorithm, we employ the transformation set $\mathcal{T}=\{\sin(\cdot), \exp(\min\{5, \cdot\}), \sqrt{|\cdot|}, \mathrm{mul}\colon(a,b)\mapsto ab\}$ and set the maximum depth $D=6$, the maximum number of nodes to be generated $N=3$, the maximum number of nodes used for generation $M=3$, and the tolerance parameter $\gamma=2$.
We considered two types of covariance matrices: $\Sigma=I_n\in\mathbb{R}^{n\times n}$ (independence) and $\Sigma=(0.5^{|i-j|})_{ij}\in\mathbb{R}^{n\times n}$ (correlation).

\begin{figure}[h]
    {
        \begin{minipage}[b]{0.49\linewidth}
            \centering
            \includegraphics[width=0.98\linewidth]{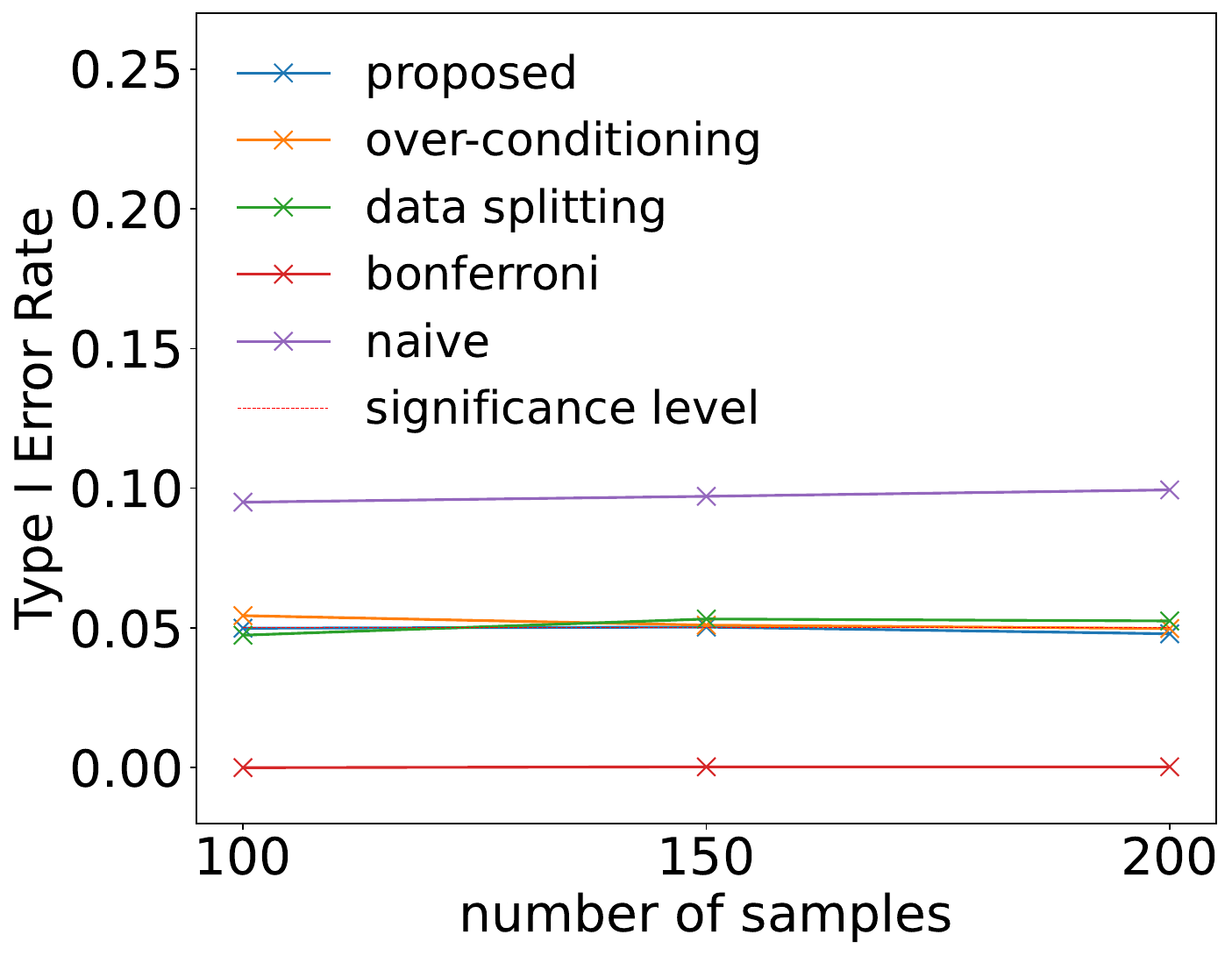}
        \end{minipage}
        \begin{minipage}[b]{0.49\linewidth}
            \centering
            \includegraphics[width=0.98\linewidth]{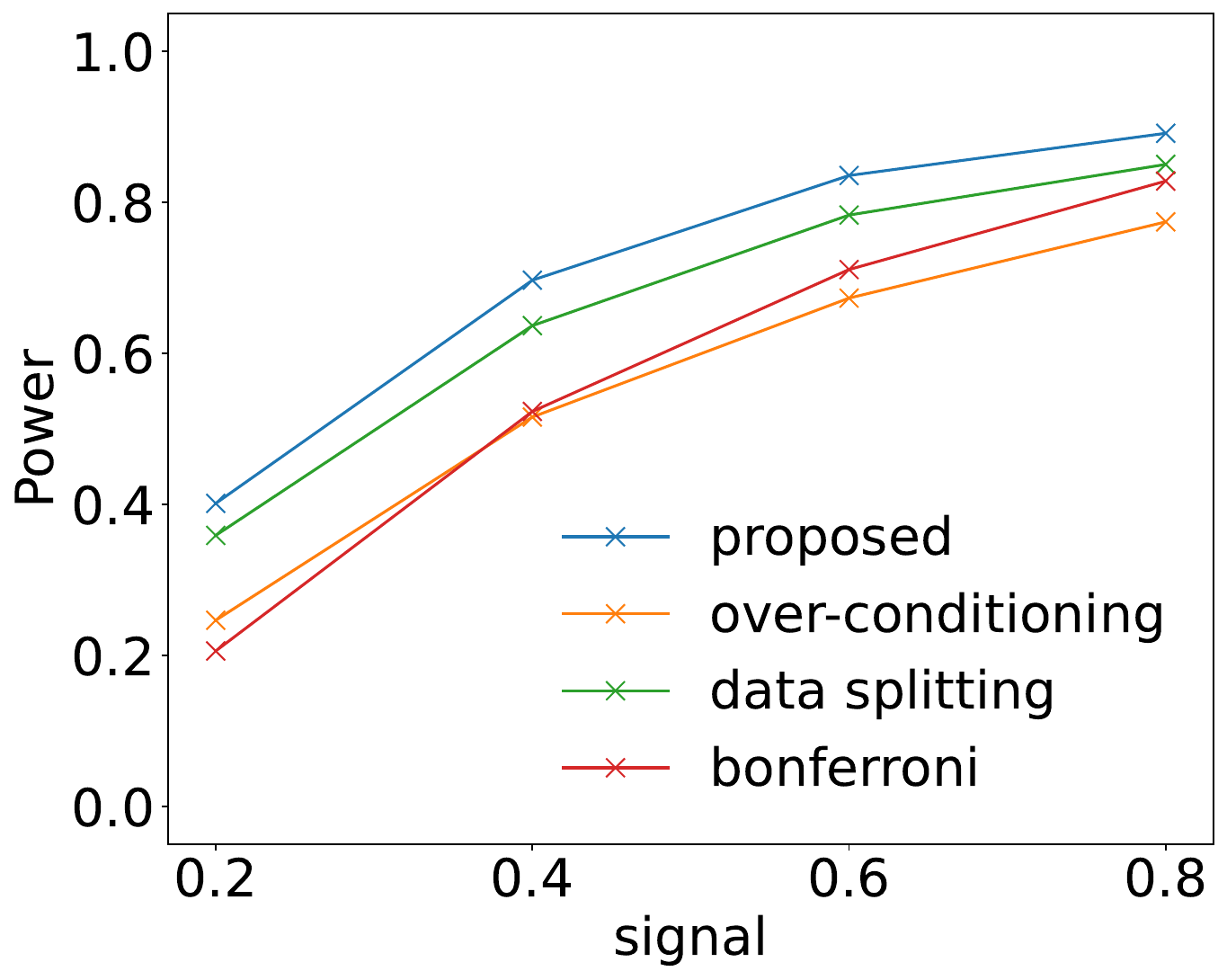}
        \end{minipage}
        \subcaption{Type I Error Rate and Power for $\Sigma=I_n$}
    }
    {
        \begin{minipage}[b]{0.49\linewidth}
            \centering
            \includegraphics[width=0.98\linewidth]{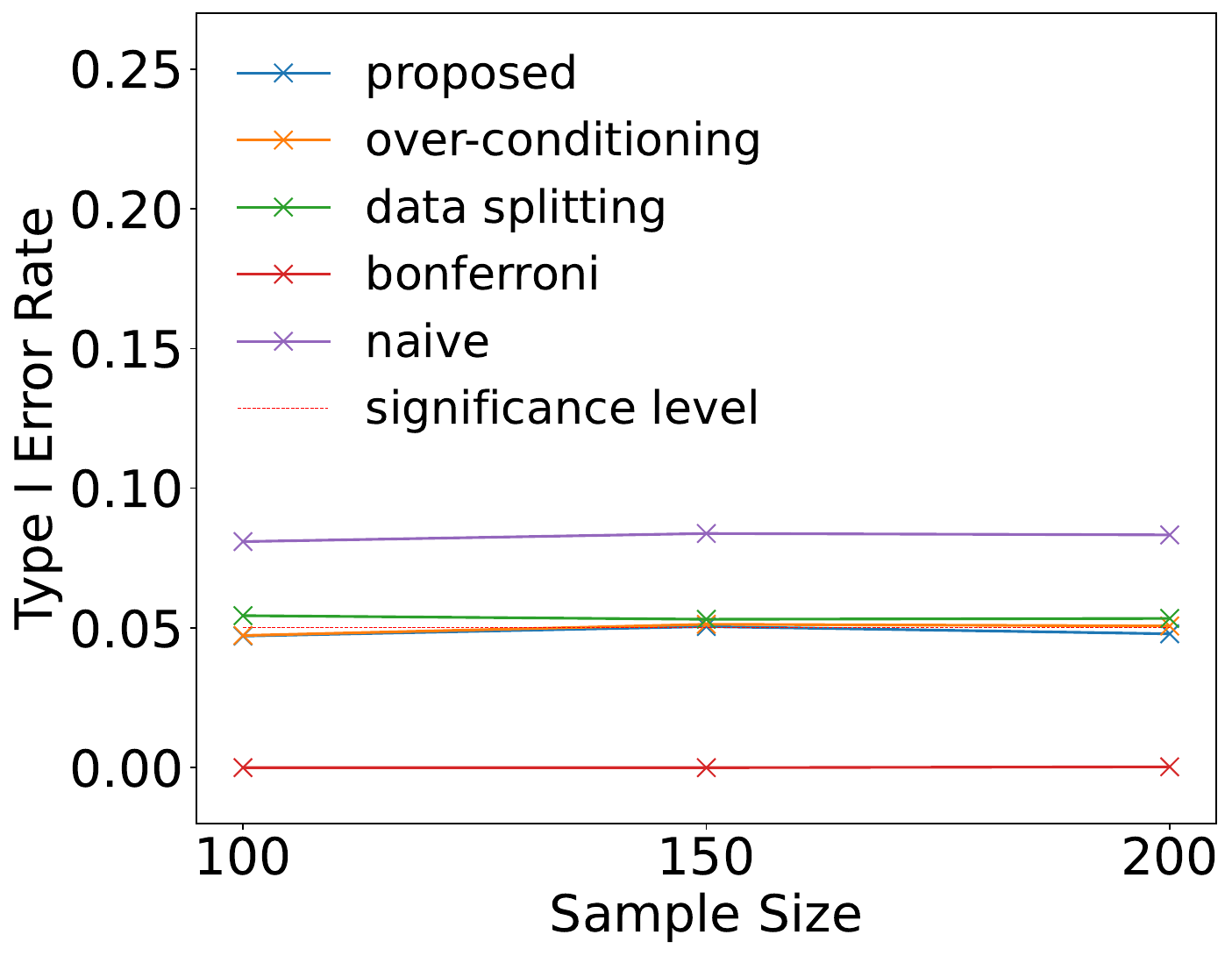}
        \end{minipage}
        \begin{minipage}[b]{0.49\linewidth}
            \centering
            \includegraphics[width=0.98\linewidth]{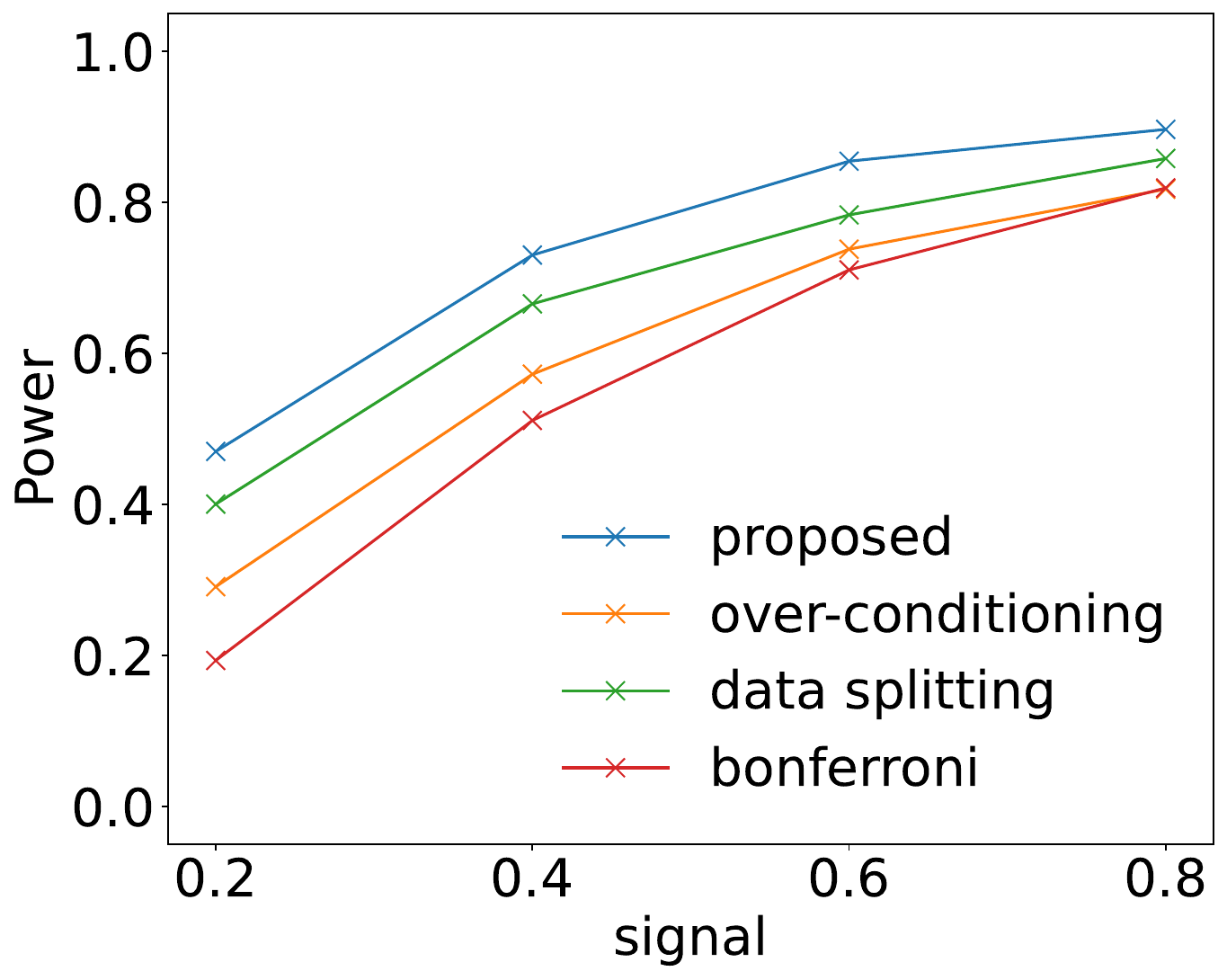}
        \end{minipage}
        \subcaption{Type I Error Rate and Power for $\Sigma=(0.5^{|i-j|})_{ij}$}
    }
    \caption{
        Type I Error Rate when changing the number of samples $n$ (left side) and Power when changing the true coefficient (right side).
        All but the naive method are able to control the type I error rate, but of these methods, our proposed method has the highest power in all settings for both covariance matrices.
        %
    }
    \label{fig:main_results}
\end{figure}

For the experiments to see the type I error rate, we change the number of samples $n\in \{100, 150, 200\}$ and set the number of features $m$ to $4$.
%
%
%
%
In each setting, we generated 10,000 null datasets $(X, \bm{y})$, where $X_{ij}\sim\mathcal{N}(0,1),\forall(i,j)\in[n]\times[m]$, $\bm{y}\sim\mathcal{N}(0, \Sigma)$. 
To investigate the power, we set $n=150$ and $m=4$ and generated dataset $(X, \bm{y})$, where $X_{ij}\sim\mathcal{N}(0,1),\forall(i,j)\in[n]\times[m]$, $\bm{y}=\Delta(\exp(\bm{x}_2) + \bm{x}_4\exp(\bm{x}_2) + \sin(\exp(\bm{x_2})) + \sqrt{|\bm{x}_4|\exp(\bm{x}_2)}) + \bm{\epsilon}$ with $\bm{\epsilon}\sim\mathcal{N}(0, I_n)$, $\Delta\in\mathbb{R}$ is an intensity of the signal.
We set $\Delta\in\{0.2,0.4,0.6,0.8\}$.
When calculating power, hypothesis testing was performed only when the true features used in dataset generation were generated by the AFE algorithm, for a total of 10,000 tests.
In addition, we provide an analysis of the computational time of the proposed method in Appendix~\ref{app:computational_time}.
See Appendix~\ref{app:computer_resources} for the computer resources used in the experiments.
\paragraph{Results.}
The results of type I error rate are shown in left side of Figure~\ref{fig:main_results}. 
The \texttt{proposed}, \texttt{oc}, \texttt{ds}, and \texttt{bonferroni} successfully controlled the type I error rate under the significance level in all settings for both covariance matrices, whereas the \texttt{naive} could not.
Because the \texttt{naive} failed to control the type I error rate, we no longer consider its power.
The results of power are shown in right side of Figure~\ref{fig:main_results} and we confirmed that the \texttt{proposed} has highest power among the methods that controlled the type I error rate, in all settings for both covariance matrices.
The \texttt{ds} has the second-highest power in all settings for both covariance matrices, but the power of the \texttt{oc} and \texttt{bonferroni} varies depending on the settings and the covariance matrices.
\paragraph{Real Data Experiments.}
%
%
We compared the \texttt{proposed} and \texttt{ds} in terms of power and AIC value on eight real datasets from the UCI Machine Learning Repository (all licensed under the CC BY 4.0; see Appendix~\ref{app:real_datasets} for more details).
%
%
From each of the original dataset, we randomly generated 10,000 sub-sampled datasets with size $n\in\{100,150,200\}$.
We applied the \texttt{proposed} and \texttt{ds} to each sub-sampled dataset.
The results are shown in Table~\ref{tab:real_data}.
In all datasets for all $n\in\{100,150,200\}$, the \texttt{proposed} has higher (better) power and lower (better) AIC value than the \texttt{ds}.
These results show that the proposed method is not only a superior test with high statistical power, but also that it does not suffer from the performance degradation of the AFE algorithm caused by the reduction in sample size inherent in the data splitting method.
%
%
\begin{table*}[ht]
    \centering
    \caption{
        Power and AIC on the real datasets when changing the number of samples.
        Each cell indicates two powers or AIC values: \texttt{proposed} v.s. \texttt{ds}, which are separated by a slash.
        For power, the higher one is better and is boldfaced; for AIC value, conversely, lower one is better and is boldfaced.
        Our proposed method has higher power and lower AIC value than the data splitting in all datasets for all $n\in\{100,150,200\}$.
        Note that the standard errors of the AIC values are all less than 0.005 and are therefore omitted from the table.
    }
    \label{tab:real_data}
    \centering
    \begin{tabular}{cccccc}
        \toprule
         & $n$   & Data1              & Data2              & Data3              & Data4              \\
        \midrule
        \multirow{3}{*}{Power}
         & $100$ & \textbf{.15}/.14   & \textbf{.23}/.20   & \textbf{.34}/.30   & \textbf{.32}/.28   \\
         & $150$ & \textbf{.20}/.19   & \textbf{.31}/.29   & \textbf{.43}/.39   & \textbf{.39}/.36   \\
         & $200$ & \textbf{.26}/.24   & \textbf{.36}/.35   & \textbf{.50}/.46   & \textbf{.46}/.43   \\
        \midrule
        \multirow{3}{*}{$\dfrac{\mathrm{AIC}}{100}$}
         & $100$ & \textbf{0.93}/0.96 & \textbf{0.81}/0.86 & \textbf{0.79}/0.84 & \textbf{0.85}/0.90 \\
         & $150$ & \textbf{1.35}/1.39 & \textbf{1.15}/1.19 & \textbf{1.08}/1.12 & \textbf{1.18}/1.21 \\
         & $200$ & \textbf{1.78}/1.81 & \textbf{1.48}/1.52 & \textbf{1.37}/1.40 & \textbf{1.50}/1.53 \\
        \bottomrule
    \end{tabular}\\
    \vspace{5mm}
    \centering
    \begin{tabular}{cccccc}
        \toprule
         & $n$   & Data5              & Data6              & Data7              & Data8              \\
        \midrule
        \multirow{3}{*}{Power}
         & $100$ & \textbf{.15}/.12   & \textbf{.14}/.12   & \textbf{.07}/.06   & \textbf{.07}/.07   \\
         & $150$ & \textbf{.20}/.18   & \textbf{.20}/.19   & \textbf{.08}/.08   & \textbf{.09}/.07   \\
         & $200$ & \textbf{.25}/.23   & \textbf{.26}/.25   & \textbf{.10}/.09   & \textbf{.10}/.09   \\
        \midrule
        \multirow{3}{*}{$\dfrac{\mathrm{AIC}}{100}$}
         & $100$ & \textbf{0.97}/1.01 & \textbf{0.96}/0.99 & \textbf{1.08}/1.11 & \textbf{1.08}/1.11 \\
         & $150$ & \textbf{1.39}/1.43 & \textbf{1.38}/1.41 & \textbf{1.56}/1.60 & \textbf{1.56}/1.59 \\
         & $200$ & \textbf{1.82}/1.87 & \textbf{1.81}/1.84 & \textbf{2.05}/2.08 & \textbf{2.04}/2.07 \\
        \bottomrule
    \end{tabular}
\end{table*}

\clearpage
\section{Conclusion}
\label{sec:conclusion}
In this paper, we developed a statistical method to evaluate the reliability of features generated by an AFE algorithm.
AFE can uncover important features that human intuition might miss, making it valuable for data-driven studies.
Since AFE algorithms rely on heuristics due to their exponential complexity, assessing their reproducibility has been challenging so far.
The proposed statistical test presented in this paper enables an unbiased evaluation of feature reproducibility based on SI framework.
While this study focuses on a simple tree search-based heuristic AFE algorithm, we believe it represents an important step toward providing theoretical guarantees for AFE.
In future work, we aim to extend this framework to offer reproducibility guarantees for various practical AFE algorithms.

\newpage
\subsection*{Acknowledgement}
This work was partially supported by MEXT KAKENHI (20H00601), JST CREST (JPMJCR21D3, JPMJCR22N2), JST Moonshot R\&D (JPMJMS2033-05), JST AIP Acceleration Research (JPMJCR21U2), NEDO (JPNP18002, JPNP20006) and RIKEN Center for Advanced Intelligence Project.

\clearpage

\appendix

\newpage
\section{Proofs}
\label{app:proofs}
\subsection{Proof of Theorem~\ref{thm:conditional_sampling_distribution}}
\label{app:proof_truncated}
According to the conditioning on $\mathcal{Q}_{\bm{Y}}=\mathcal{Q}_{\bm{y}}$, we have
\begin{equation}
    \mathcal{Q}_{\bm{Y}} = \mathcal{Q}_{\bm{y}} \Leftrightarrow
    \left(
    I_{n} -
    \frac{\Sigma\bm{\eta}\bm{\eta}^\top}{\bm{\eta}^\top\Sigma\bm{\eta}}
    \right)
    \bm{Y} =
    \mathcal{Q}_{\bm{y}}
    \Leftrightarrow
    \bm{Y} = \bm{a} + \bm{b}z,
\end{equation}
where $z=T(\bm{Y})=\bm{\eta}^\top\bm{Y}\in\mathbb{R}$. Then, we have
\begin{align}
      &
    \{
    \bm{Y}\in\mathbb{R}^{n}\mid
    \mathcal{F}_{\bm{Y}} = \mathcal{F}_{\bm{y}},
    \mathcal{Q}_{\bm{Y}} = \mathcal{Q}_{\bm{y}}
    \}  \\
    = &
    \{
    \bm{Y}\in\mathbb{R}^{n}\mid
    \mathcal{F}_{\bm{Y}} = \mathcal{F}_{\bm{y}},
    \bm{Y} = \bm{a} + \bm{b}z, z\in\mathbb{R}
    \}  \\
    = &
    \{
    \bm{a} + \bm{b}z\in\mathbb{R}^{n}\mid
    \mathcal{F}_{\bm{a}+\bm{b}z} = \mathcal{F}_{\bm{y}},
    z\in\mathbb{R}
    \}  \\
    = &
    \{
    \bm{a} + \bm{b}z\in\mathbb{R}^{n}\mid
    z\in \mathcal{Z}
    \}.
\end{align}
Therefore, we obtain
\begin{equation}
    T(\bm{Y}) \mid
    \{
    \mathcal{F}_{\bm{Y}} = \mathcal{F}_{\bm{y}},
    \mathcal{Q}_{\bm{Y}} = \mathcal{Q}_{\bm{y}}
    \}
    \sim
    \mathrm{TN}(\bm{\eta}^\top\bm{\mu}, \bm{\eta}^\top\Sigma\bm{\eta}, \mathcal{Z}).
\end{equation}
\subsection{Proof of Theorem~\ref{thm:property_of_selective_p_value}}
\label{app:proof_property_of_selective_p_value}
By probability integral transformation, under the null hypothesis, we have
\begin{equation}
    p_\mathrm{selective} \mid
    \{
    \mathcal{F}_{\bm{Y}} = \mathcal{F}_{\bm{y}},
    \mathcal{Q}_{\bm{Y}} = \mathcal{Q}_{\bm{y}}
    \}
    \sim
    \mathrm{Unif}(0, 1),
\end{equation}
which leads to
\begin{equation}
    \mathbb{P}_{\mathrm{H}_0}
    \left(
    p_\mathrm{selective} \leq \alpha \mid
    \mathcal{F}_{\bm{Y}} = \mathcal{F}_{\bm{y}},
    \mathcal{Q}_{\bm{Y}} = \mathcal{Q}_{\bm{y}}
    \right)
    =\alpha,\
    \forall\alpha\in(0,1).
\end{equation}
For any $\alpha\in(0,1)$, by marginalizing over all the values of the nuisance parameters, we obtain
\begin{align}
      &
    \mathbb{P}_{\mathrm{H}_0}
    \left(
    p_\mathrm{selective} \leq \alpha \mid
    \mathcal{F}_{\bm{Y}} = \mathcal{F}_{\bm{y}},
    \right)                                                                                                       \\
    = &
    \begin{multlined}
        \int_{\mathbb{R}^{n}}
        \mathbb{P}_{\mathrm{H}_0}
        \left(
        p_\mathrm{selective} \leq \alpha \mid
        \mathcal{F}_{\bm{Y}} = \mathcal{F}_{\bm{y}},
        \mathcal{Q}_{\bm{Y}} = \mathcal{Q}_{\bm{y}}
        \right) \\
        \mathbb{P}_{\mathrm{H}_0}
        \left(
        \mathcal{Q}_{\bm{Y}} = \mathcal{Q}_{\bm{y}} \mid
        \mathcal{F}_{\bm{Y}} = \mathcal{F}_{\bm{y}},
        \right)
        d\mathcal{Q}_{\bm{y}}
    \end{multlined} \\
    = & \alpha \int_{\mathbb{R}^{n}}
    \mathbb{P}_{\mathrm{H}_0}
    \left(
    \mathcal{Q}_{\bm{Y}} = \mathcal{Q}_{\bm{y}} \mid
    \mathcal{F}_{\bm{Y}} = \mathcal{F}_{\bm{y}},
    \right)
    d\mathcal{Q}_{\bm{y}} = \alpha.
\end{align}
Therefore, we also obtain
\begin{align}
    \mathbb{P}_{\mathrm{H}_0}(p_{\mathrm{selective}}\leq \alpha)
    = &
    \sum_{\mathcal{F}_{\bm{y}}\in \mathcal{F}(X)}
    \mathbb{P}_{\mathrm{H}_0}(\mathcal{F}_{\bm{y}})
    \mathbb{P}_{\mathrm{H}_0}
    \left(
    p_\mathrm{selective} \leq \alpha \mid
    \mathcal{F}_{\bm{Y}} = \mathcal{F}_{\bm{y}}
    \right) \\
    = &
    \alpha
    \sum_{\mathcal{F}_{\bm{y}}\in \mathcal{F}(X)}
    \mathbb{P}_{\mathrm{H}_0}(\mathcal{F}_{\bm{y}}) = \alpha.
\end{align}
\newpage
\section{Details of the Experiments}
\subsection{Methods for Comparison}
\label{app:methods_for_comparison}
We compared our proposed method with the following methods:
\begin{itemize}
    \item \texttt{oc}: Our proposed method conditioning on the only one intervals $[L_z, U_z]$ to which the observed test statistic $T(\bm{y})$ belongs. This method is computationally efficient, however, its power is low due to over-conditioning.
    \item \texttt{naive}: This method uses a classical $z$-test without conditioning, i.e., we compute the naive $p$-value as $p_\mathrm{naive}=\mathbb{P}_{\mathrm{H}_0}(|T(\bm{Y})|\geq |T(\bm{y})|)$.
    \item \texttt{bonferroni}:  This is a method to control the type I error rate by using the Bonferroni correction. The number of all possible sets of generated features is $\sum_{d\in[D]}N^d$, then we compute the bonferroni p-value as $p_\mathrm{bonferroni} = \min(1, \sum_{d\in[D]}N^d \cdot p_\mathrm{naive})$.
    \item \texttt{ds}: This method split the dataset into two parts, one for auto feature generation and the other for testing.
\end{itemize}
\subsection{Computational Time of the Proposed Method}
\label{app:computational_time}
\begin{figure}[h]
    \begin{minipage}[b]{0.24\linewidth}
        \centering
        \includegraphics[width=0.98\linewidth]{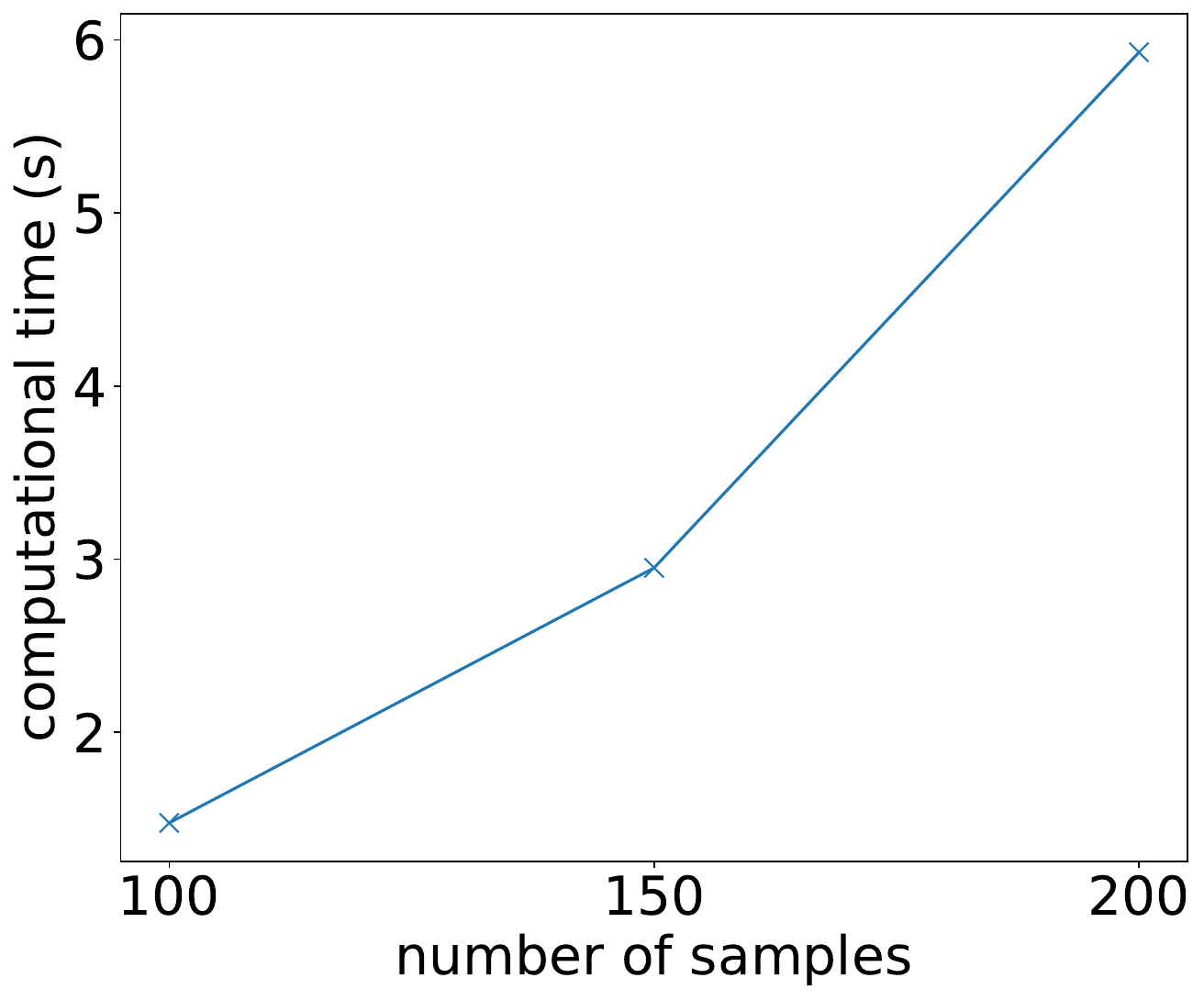}
        \subcaption{$\Sigma=I_n$}
    \end{minipage}
    \begin{minipage}[b]{0.24\linewidth}
        \centering
        \includegraphics[width=0.98\linewidth]{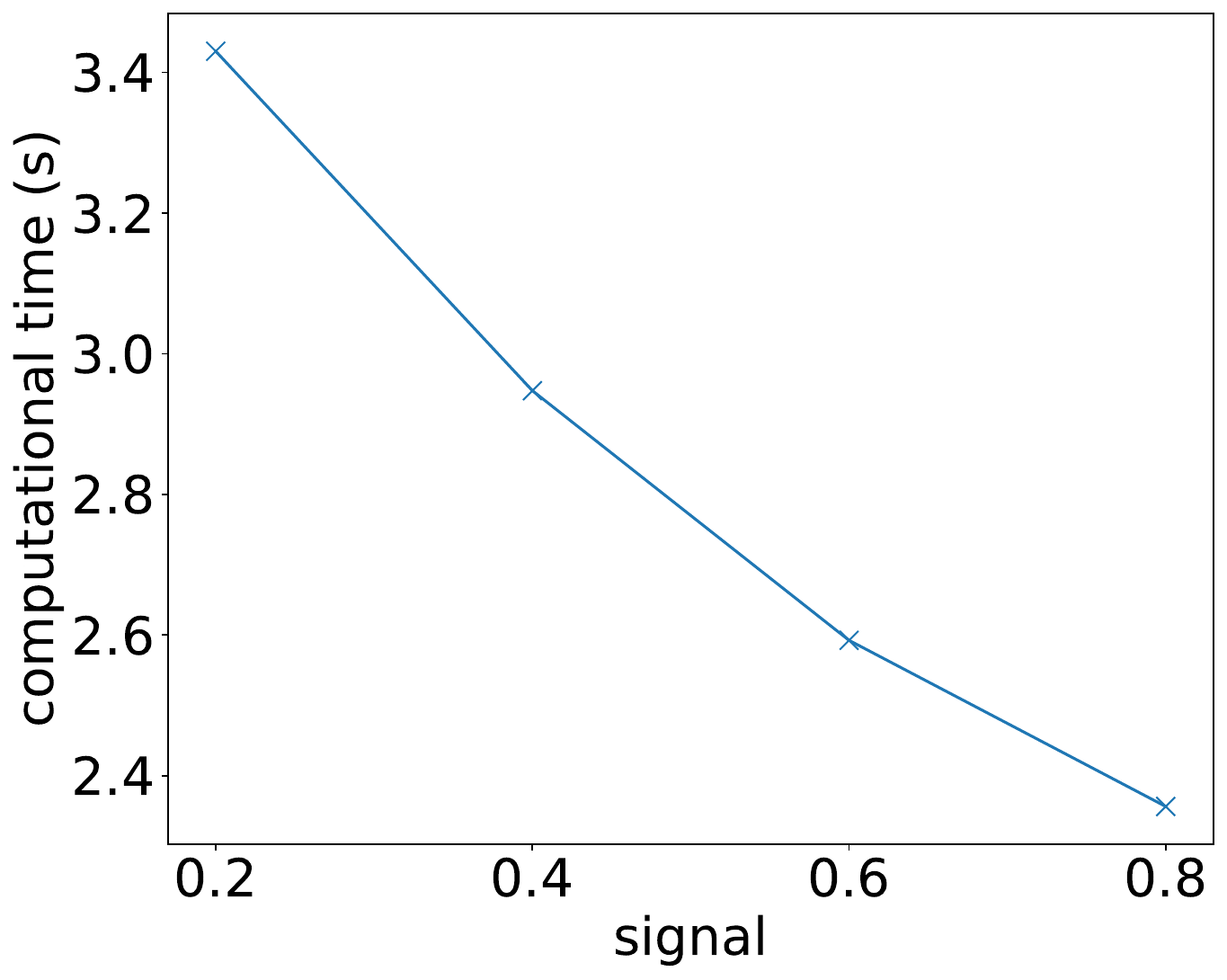}
        \subcaption{$\Sigma=I_n$}
    \end{minipage}
    \begin{minipage}[b]{0.24\linewidth}
        \centering
        \includegraphics[width=0.98\linewidth]{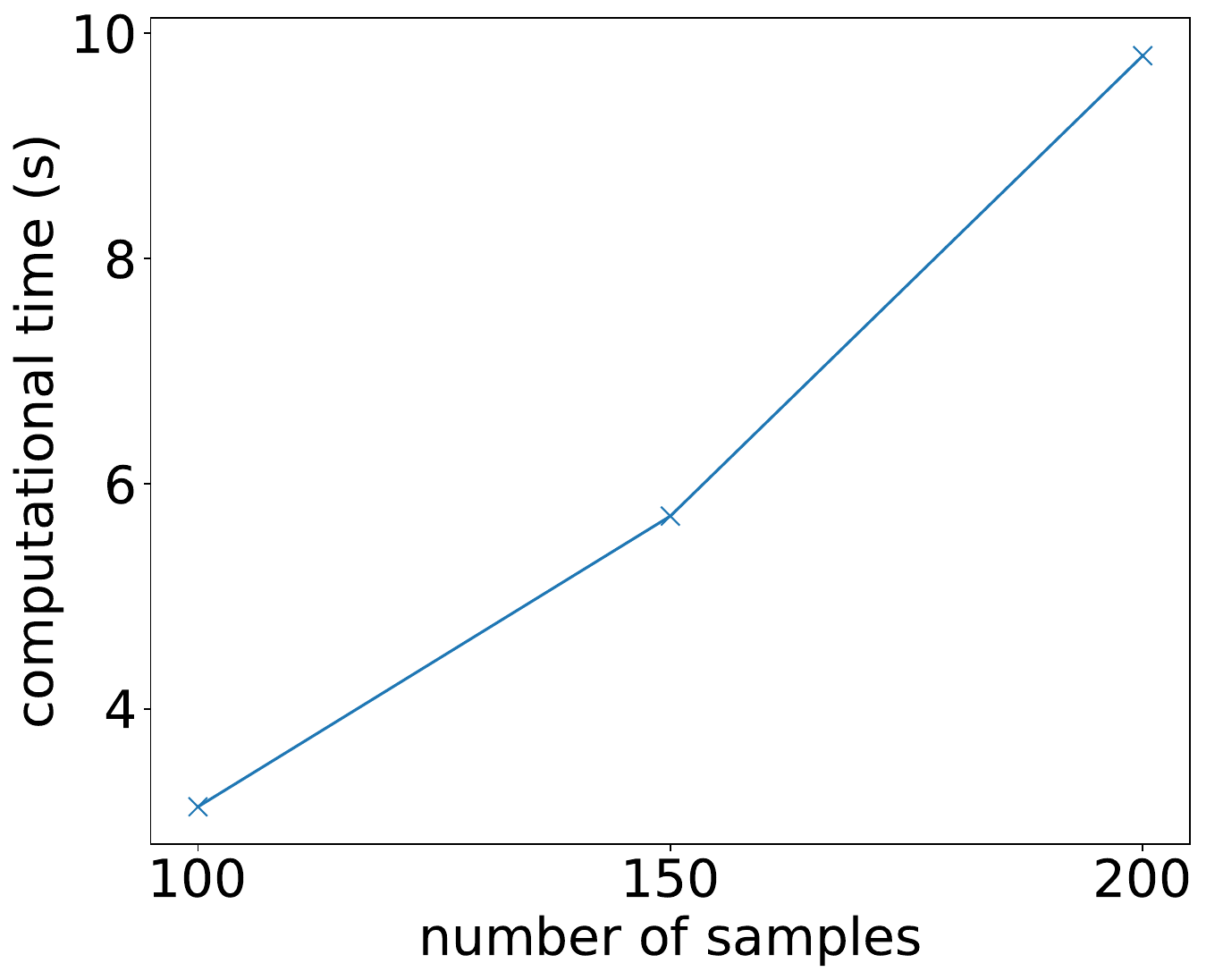}
        \subcaption{$\Sigma=(0.5^{|i-j|})_{ij}$}
    \end{minipage}
    \begin{minipage}[b]{0.24\linewidth}
        \centering
        \includegraphics[width=0.98\linewidth]{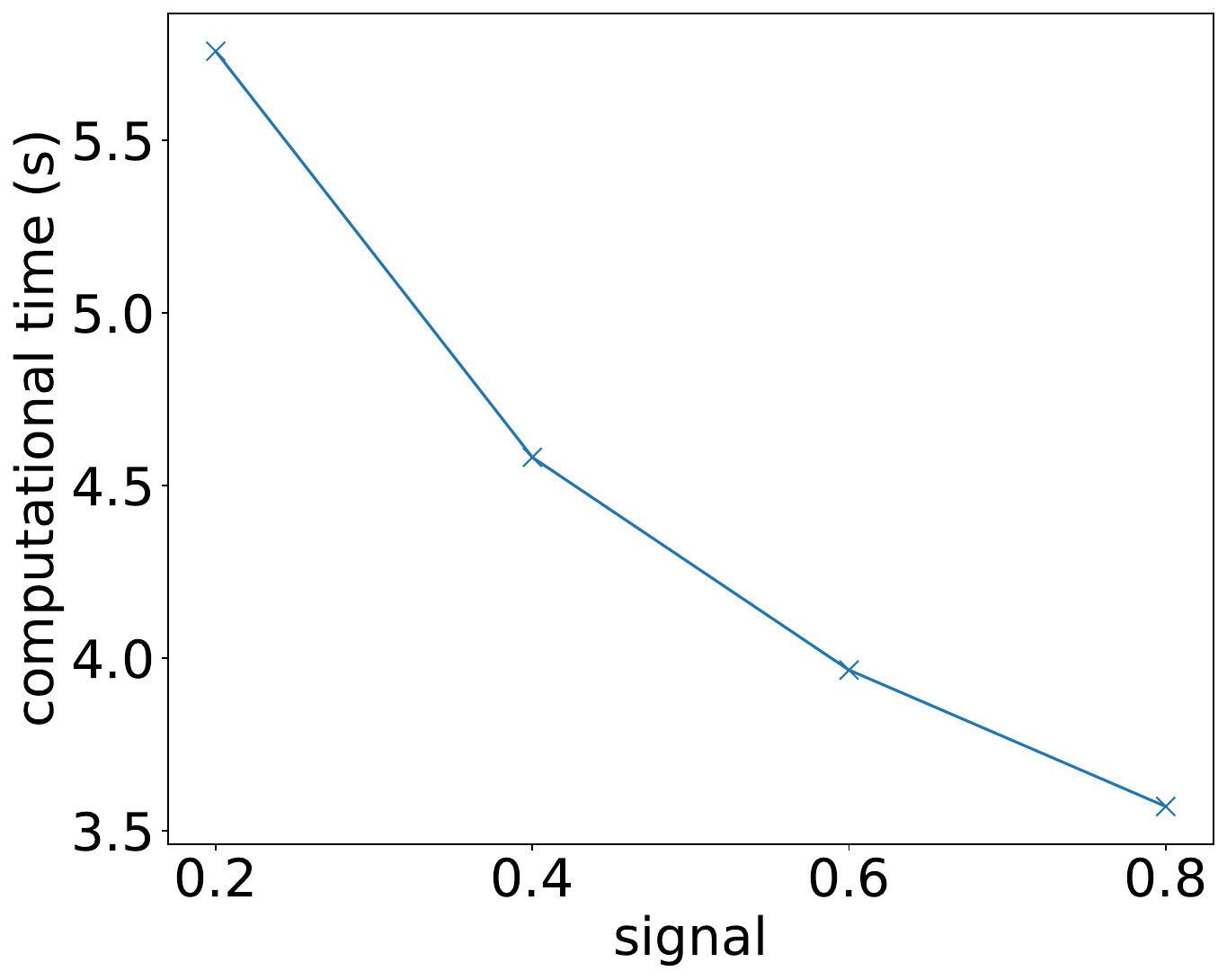}
        \subcaption{$\Sigma=(0.5^{|i-j|})_{ij}$}
    \end{minipage}
    \caption{
        Computational Time when evaluating type I error rate (left side) and Power (right side).
        The results show that computational time is exponentially increased as the number of samples increases.
        In addition, the higher the signal intensity, the shorter the computational time, which may be due to the fact that the results of hypothesis testing are more likely to be obvious.
    }
    \label{fig:time_comparison}
\end{figure}

We analyzed the computational time of our proposed method in the run of our experiments on synthetic datasets in \S\ref{sec:experiments}.
The results are shown in Figure~\ref{fig:time_comparison}.
\subsection{Computer Resources}
\label{app:computer_resources}
All numerical experiments were conducted on a computer with a 96-core 3.60GHz CPU and 512GB of memory.
\subsection{Details of the Real Datasets}
\label{app:real_datasets}
We used the following eight real datasets from the UCI Machine Learning Repository.
All datasets are licensed under the CC BY 4.0 license.
\begin{itemize}
    \item Airfoil Self-Noise~\citep{airfoil_self-noise_291} for Data1
    \item Concrete Compressive Strength~\citep{concrete_compressive_strength_165} for Data2
    \item Energy Efficiency~\citep{energy_efficiency_242} for Data3 (heating load) and Data4 (cooling load)
    \item Gas Turbine CO and NOx Emission Data Set~\citep{gas_turbine_co_and_nox_emission_data_set_551} for Data5
    \item Real Estate Valuation~\citep{real_estate_valuation_477} for Data6
    \item Wine Quality~\citep{wine_quality_186} for Data7 (red wine) and Data8 (white wine)
\end{itemize}

\newpage
\section{Robustness of Type I Error Rate Control}
\label{app:robustness}
In this experiment, we confirmed the robustness of the proposed method in terms of type I error rate control by applying our method to the two cases: the case where the variance is estimated from the same data and the case where the noise is non-Gaussian.
\subsection{Estimated Variance}
In the case where the variance is estimated from the same data, we considered the same setting as in type I error rate experiments in~\S\ref{sec:experiments}.
For each setting, we generated 10,000 null datasets $(X, \bm{y})$, where $X_{ij}\sim\mathcal{N}(0,1),\ \forall(i,j)\in[n]\times[m]$ and $\bm{y}\sim\mathcal{N}(0, I_n)$ and estimated the variance $\hat{\sigma}^2$ as
\begin{equation}
    \hat{\sigma}^2 = \frac{1}{n-m}\|\bm{y}-X(X^\top X)^{-1}X^\top\bm{y}\|_2^2.
\end{equation}
We considered the three significance levels $\alpha=0.05,0.01,0.10$.
The result is shown in Figure~\ref{fig:estimated_variance} and our proposed method can properly control the type I error rate.
\subsection{Non-Gaussian Noise}
In the case where the noise is non-Gaussian, we set $n=150$ and $m=4$.
As non-Gaussian noise, we considered the following five distribution families:
\begin{itemize}
    \item \texttt{skewnorm}: Skew normal distribution family.
    \item \texttt{exponnorm}: Exponentially modified normal distribution family.
    \item \texttt{gennormsteep}: Generalized normal distribution family (limit the shape parameter $\beta$ to be steeper than the normal distribution, i.e., $\beta < 2$).
    \item \texttt{gennormflat}: Generalized normal distribution family (limit the shape parameter $\beta$ to be flatter than the normal distribution, i.e., $\beta > 2$).
    \item \texttt{t}: Student's t distribution family.
\end{itemize}
Note that all of these distribution families include the Gaussian distribution and are standardized in the experiment.

\begin{figure}[ht]
    \begin{minipage}[b]{0.49\linewidth}
        \centering
        \includegraphics[width=0.7\linewidth]{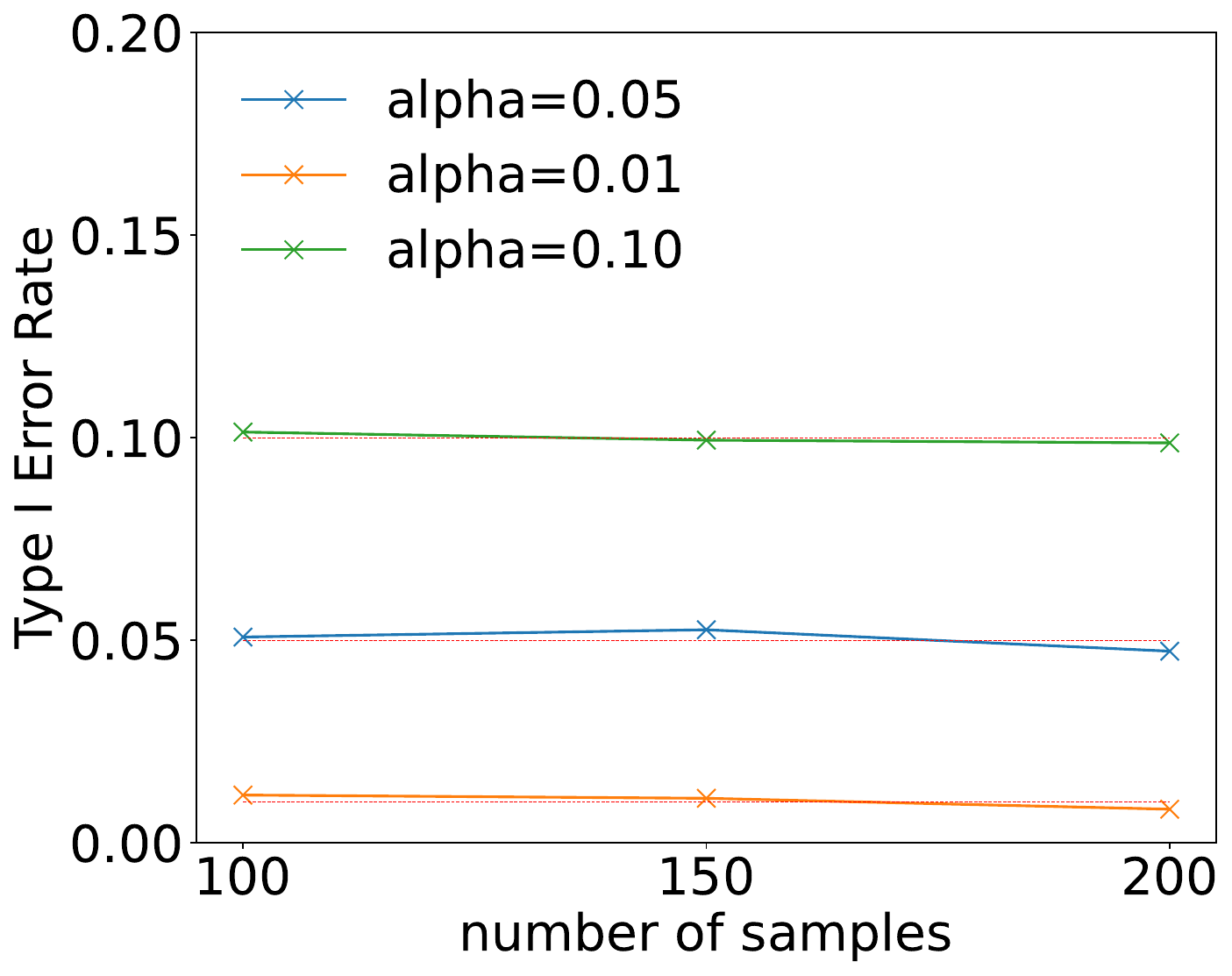}
        \subcaption{Estimated Variance}
        \label{fig:estimated_variance}
    \end{minipage}
    \begin{minipage}[b]{0.49\linewidth}
        \centering
        \includegraphics[width=0.7\linewidth]{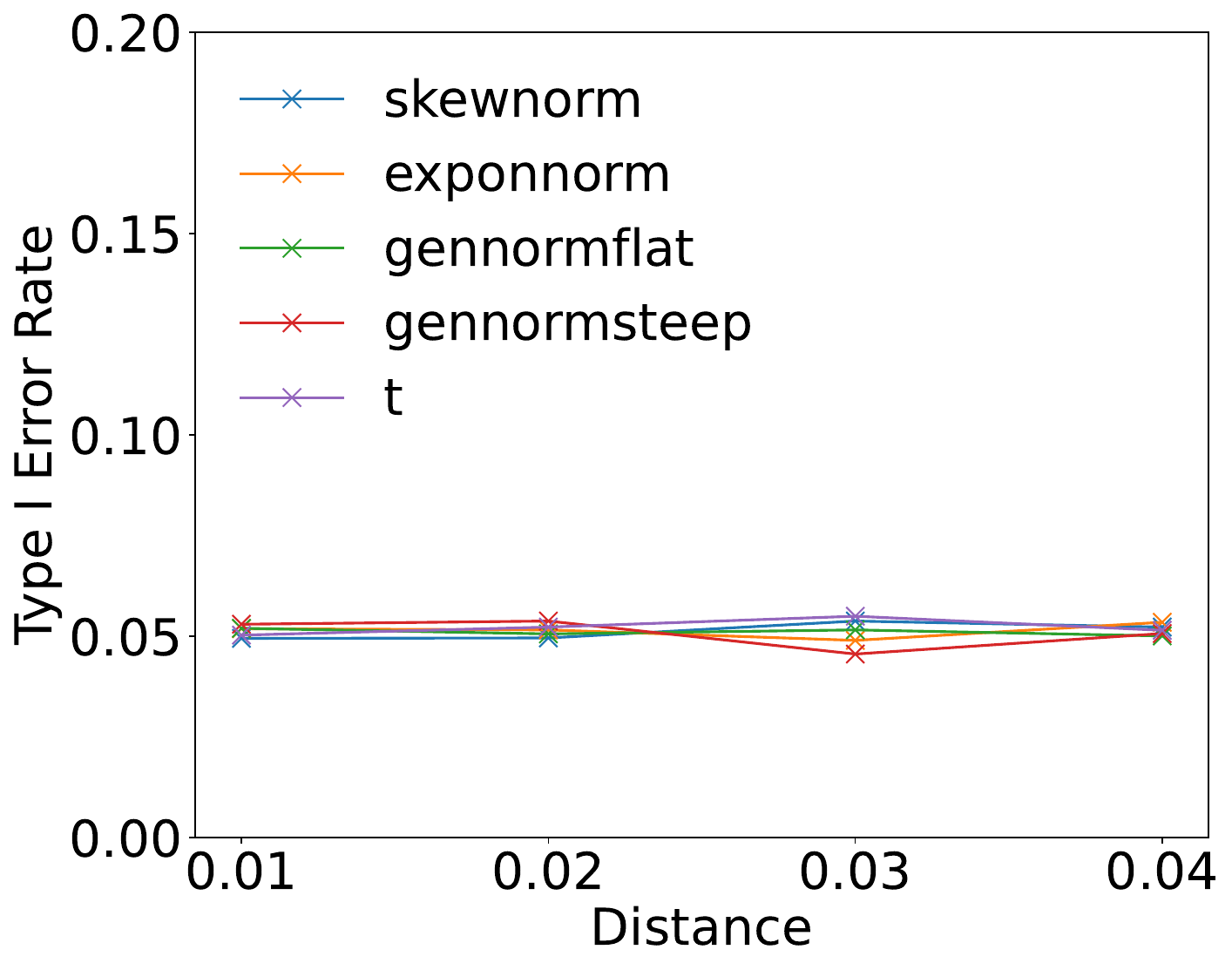}
        \subcaption{Non-Gaussian Noise}
        \label{fig:non_gaussian}
    \end{minipage}
    \caption{
        Robustness of Type I Error Rate Control.
        Our proposed method can robustly control the type I error rate even for two cases: where the variance is estimated from the same data and where the noise is non-Gaussian.
    }
\end{figure}

To conduct the experiment, we first obtained a distribution such that the 1-Wasserstein distance from $\mathcal{N}(0,1)$ is $l$ in each distribution family, for $l\in\{0.01,0.02,0.03,0.04\}$.
We then generated 10,000 null datasets $(X, \bm{y})$, where $X_{ij}\sim\mathcal{N}(0,1),\ \forall(i,j)\in[n]\times[m]$ and $\bm{y}_i,\ \forall i\in[n]$ follows the obtained distribution.
We considered the significance level $\alpha=0.05$.
The result is shown in Figure~\ref{fig:non_gaussian} and our proposed method can properly control the type I error rate.
\clearpage

\bibliographystyle{plainnat}
\bibliography{ref}

\end{document}